\documentclass[11pt,letterpaper]{article}
\usepackage[margin=1in]{geometry}
\usepackage[T1]{fontenc}
\usepackage{mathpazo}
\usepackage{amsmath,amssymb,amsthm,mathtools}
\usepackage{dsfont,nicefrac}
\usepackage{thm-restate}
\usepackage{microtype,enumitem,needspace,array,etoolbox}
\usepackage{xcolor}
\definecolor{blueGrotto}{HTML}{059DC0}
\definecolor{royalBlue}{HTML}{057DCD}
\definecolor{navyBlue}{HTML}{0B579C}
\usepackage{hyperref}
\hypersetup{colorlinks=true,urlcolor=blueGrotto,linkcolor=royalBlue,citecolor=navyBlue}
\usepackage[capitalise,noabbrev,nameinlink,sort]{cleveref}
\usepackage[
  backref=true,backend=biber,natbib=true,style=alphabetic,
  sorting=alphabeticlabel,sortcites=true,
  minbibnames=3,maxbibnames=999,mincitenames=4,maxcitenames=4,
  minalphanames=4,maxalphanames=4,doi=false,isbn=false
]{biblatex}
\DeclareSortingTemplate{alphabeticlabel}{
  \sort[final]{\field{labelalpha}}
  \sort{\field{year}}
  \sort{\field{title}}
}
\AtEveryBibitem{\clearname{editor}\clearlist{location}}

\allowdisplaybreaks[2]
\setlist[enumerate]{label=(\roman*),leftmargin=2.2em}
\numberwithin{equation}{section}

\theoremstyle{plain}
\newtheorem{theorem}{Theorem}[section]
\newtheorem{informaltheorem}{Informal Theorem}
\newtheorem{lemma}{Lemma}
\newtheorem{proposition}{Proposition}
\newtheorem{corollary}{Corollary}
\newtheorem{claim}{Claim}
\newtheorem{assumption}{Assumption}
\theoremstyle{definition}
\newtheorem{remark}{Remark}

\newtheorem{problem}{Problem}
\newtheorem{introdefinition}{Definition}

\makeatletter
\@for\codex@sharedtheorem:=informaltheorem,lemma,proposition,corollary,remark,definition\do{%
  \expandafter\let\csname c@\codex@sharedtheorem\endcsname\c@theorem
  \expandafter\let\csname the\codex@sharedtheorem\endcsname\thetheorem
  \expandafter\let\csname p@\codex@sharedtheorem\endcsname\p@theorem
  \expandafter\let\csname theH\codex@sharedtheorem\endcsname\theHtheorem
}
\makeatother

\crefname{informaltheorem}{Informal Theorem}{Informal Theorems}
\Crefname{informaltheorem}{Informal Theorem}{Informal Theorems}
\crefname{lemma}{Lemma}{Lemmas}
\crefname{proposition}{Proposition}{Propositions}
\crefname{corollary}{Corollary}{Corollaries}
\crefname{remark}{Remark}{Remarks}
\crefname{definition}{Definition}{Definitions}
\crefname{problem}{Problem}{Problems}
\crefname{introdefinition}{Definition}{Definitions}

\crefname{claim}{Claim}{Claims}
\Crefname{claim}{Claim}{Claims}
\crefname{assumption}{Assumption}{Assumptions}
\Crefname{assumption}{Assumption}{Assumptions}
\crefname{example}{Example}{Examples}
\Crefname{example}{Example}{Examples}

\AfterEndEnvironment{theorem}{\noindent\ignorespaces}
\AfterEndEnvironment{informaltheorem}{\noindent\ignorespaces}
\AfterEndEnvironment{lemma}{\noindent\ignorespaces}
\AfterEndEnvironment{proposition}{\noindent\ignorespaces}
\AfterEndEnvironment{corollary}{\noindent\ignorespaces}
\AfterEndEnvironment{claim}{\noindent\ignorespaces}
\AfterEndEnvironment{assumption}{\noindent\ignorespaces}
\AfterEndEnvironment{remark}{\noindent\ignorespaces}
\AfterEndEnvironment{definition}{\noindent\ignorespaces}
\AfterEndEnvironment{example}{\noindent\ignorespaces}
\AfterEndEnvironment{problem}{\noindent\ignorespaces}
\AfterEndEnvironment{introdefinition}{\noindent\ignorespaces}
\AfterEndEnvironment{proof}{\noindent\ignorespaces}
\AfterEndEnvironment{proofof}{\noindent\ignorespaces}
\AfterEndEnvironment{restatable}{\noindent\ignorespaces}

\DeclareMathOperator{\Unif}{\operatorname{Unif}}

\newcommand{\cX}{\mathcal X}
\newcommand{\cH}{\mathcal H}
\newcommand{\cD}{\mathcal D}
\newcommand{\cB}{\mathcal B}

\newcommand{\Nat}{\mathbb N}
\newcommand{\Nzero}{\mathbb N_0}
\newcommand{\R}{\mathbb R}
\newcommand{\E}{\mathbb E}
\newcommand{\1}{\mathds{1}}
\newcommand{\CLOS}{\operatorname{CLOS}}
\newcommand{\VCdim}{\operatorname{VCdim}}
\newcommand{\supp}{\operatorname{supp}}
\newcommand{\err}{\operatorname{err}}

\newcommand{\cc}{\mathfrak c}

\newcommand{\cP}{\mathcal P}

\providecommand{\inbrace}[1]{\left\{#1\right\}}
\providecommand{\inparen}[1]{\left(#1\right)}
\providecommand{\insquare}[1]{\left[#1\right]}
\providecommand{\abs}[1]{\left|#1\right|}
\providecommand{\ceil}[1]{\left\lceil#1\right\rceil}

\usepackage{color-edits}
\addauthor[Shai]{sbd}{green}
\addauthor[Farnam]{fm}{blue}
\addauthor[Anay]{am}{cyan}
\definecolor{anayAI}{rgb}{0.50,0.10,0.60}
\addauthor[anay-AI]{amai}{anayAI}
\addauthor[farnam-AI]{fmai}{teal}

\makeatletter
\ifthenelse{\equal{\coloredits@SuppressEdits}{no}}{%
  \renewcommand{\amaiedit}[1]{%
    \ifmmode\else\leavevmode\fi{\color{anayAI}#1}}
  \renewcommand{\amaicomment}[1]{%
    \ifmmode\else\leavevmode\fi{\color{anayAI}[anay-AI: #1]}}
  
}{}
\makeatother

\makeatletter
\ifthenelse{\equal{\coloredits@SuppressEdits}{no}}{%
  \renewcommand{\amedit}[1]{%
    \ifmmode\else\leavevmode\fi{\color{cyan}#1}}
  
  \renewcommand{\amdelete}[1]{%
    \ifmmode\else\leavevmode\fi{\color{cyan}$[\bullet]$}%
    \marginpar{\parbox[t]{0.83in}{\raggedright\scriptsize\color{cyan}Anay deleted here}}}
  \renewcommand{\ammargincomment}[1]{%
    \ifmmode\else\leavevmode\fi{\color{cyan}$[\bullet]$}%
    \marginpar{\parbox[t]{0.83in}{\raggedright\scriptsize\color{cyan}[Anay: #1]}}}
}{}
\makeatother

\definecolor{anayClaude}{RGB}{235,110,0}
\addauthor[anay-claude]{ac}{anayClaude}
\makeatletter
\ifthenelse{\equal{\coloredits@SuppressEdits}{no}}{%
  \renewcommand{\acedit}[1]{%
    \ifmmode\else\leavevmode\fi{\color{anayClaude}#1}}
  \renewcommand{\accomment}[1]{%
    \ifmmode\else\leavevmode\fi{\color{anayClaude}[anay-claude: #1]}}
  \renewcommand{\acdelete}[1]{}
  
  \renewcommand{\acmargincomment}[1]{%
    \ifmmode\else\leavevmode\fi{\color{anayClaude}$[\bullet]$}%
    \marginpar{\raggedright\scriptsize\color{anayClaude}[anay-claude: #1]}}
}{}
\makeatother

\usepackage[normalem]{ulem}
\definecolor{anayClaudeRed}{RGB}{200,0,0}
\definecolor{anayClaudeOld}{gray}{0.45}
\addauthor[anay-claude]{acmajor}{anayClaudeRed}
\makeatletter
\ifthenelse{\equal{\coloredits@SuppressEdits}{no}}{%
  \renewcommand{\acmajoredit}[1]{%
    \ifmmode\else\leavevmode\fi{\color{anayClaudeRed}#1}}
  \renewcommand{\acmajordelete}[1]{%
    \ifmmode\else\leavevmode\fi{\color{anayClaudeOld}\sout{#1}}}
  
  \renewcommand{\acmajormargincomment}[1]{%
    \ifmmode\else\leavevmode\fi{\color{anayClaudeRed}$[\bullet]$}%
    \marginpar{\parbox[t]{0.83in}{\raggedright\scriptsize\color{anayClaudeRed}[anay-claude: #1]}}}
}{}
\makeatother

\title{Deterministic Proper Positive-Only\\Learning Is Independent of ZFC}

\hypersetup{pdftitle={Deterministic Proper Positive-Only Learning Is Independent of ZFC},pdfauthor={Shai Ben-David, Farnam Mansouri, Anay Mehrotra}}

\definecolor{codexColor}{RGB}{105,45,150}
\addauthor[Codex]{codex}{codexColor}
\makeatletter
\ifthenelse{\equal{\coloredits@SuppressEdits}{no}}{%
}{}
\makeatother
\newcolumntype{C}[1]{>{\centering\let\newline\\\arraybackslash\hspace{0pt}}m{#1}}
\author{
  \begin{tabular}{C{.295\textwidth}C{.295\textwidth}C{.295\textwidth}}
    \textbf{Shai Ben-David} & \textbf{Farnam Mansouri} & \textbf{Anay Mehrotra}\\
    \small University of Waterloo & \small University of Waterloo & \small Stanford University\\
    \small Vector Institute & \small Vector Institute & \\
    \small\mbox{\href{mailto:shai@uwaterloo.ca}{shai@uwaterloo.ca}} &
    \small\mbox{\href{mailto:f5mansou@uwaterloo.ca}{f5mansou@uwaterloo.ca}} &
    \small\mbox{\href{mailto:anaymehrotra1@gmail.com}{anaymehrotra1@gmail.com}}
  \end{tabular}
}
\date{}

\begin{document}
\maketitle

\begin{abstract}
Binary classification from positive-only samples is a variant of PAC
learning in which the learner receives i.i.d.\ samples only from the positive
region of an unknown target concept, but is evaluated on the full
distribution. Improper learnability in this model has a textbook
characterization, whereas proper learnability has remained open for decades.
Recent work characterized
it for learners with access to randomness, and
showed that, unlike in PAC learning, some classes are properly learnable with randomness but not without it, leaving the deterministic
case open. 
We show that this case
cannot be resolved within the standard axioms of mathematics (ZFC): there is a class of
VC dimension two whose deterministic proper learnability is independent of
ZFC, even though a randomized learner provably learns it. Moreover,
no dimension of finite character, a desirable property shared by most
dimensions in learning theory, characterizes
deterministic proper learnability. We also show that the size of the
domain is decisive. In ZFC, deterministic
proper learnability has a combinatorial characterization on
every domain whose cardinality is below a certain threshold, including every
countable domain. In contrast, this characterization fails
on every domain of sufficiently large cardinality. Under the generalized continuum hypothesis, this threshold is sharp: it is the exact cardinality at which the
characterization starts to fail. Finally, our proofs introduce new tools, based on the
Erd\H{o}s--Rado partition theorem, that may be of broader interest.
\end{abstract}

\clearpage
\tableofcontents
\clearpage

\section{Introduction}\label{sec:intro}
\vspace{-7pt}

\begingroup
\linespread{1.06}\selectfont
In the foundational model of PAC learning \citep{Valiant1984}, a learner observes examples drawn from an
unknown distribution and labeled by an unknown target concept, and must output a
classifier with small misclassification error. The model implicitly
assumes access to both positive and negative examples. In practice, however, negative
examples are often expensive, scarce, or simply unavailable.
This issue was already noted by
Valiant, who wrote: ``While it may be
reasonable to discuss the distribution of the attributes of elephants, we
may prefer not discussing the distribution of the attributes of
non-elephants'' \citep{Valiant1984b}. It also arises in many modern
applications. In medical
diagnosis, records list
confirmed cases of a condition, but they rarely certify that a patient does
not have it \citep{ElkanNoto2008}. In recommendation
systems, an interaction provides positive
evidence, whereas a missing interaction need not indicate a negative
preference \citep{LiuEtAl2003}. Positive-only learning, a natural
variant of the PAC model, asks what can be
learned when these positive observations are the learner's entire source
of data.

Formally, examples are drawn from an unknown distribution $\cD$ over a
domain $\cX$ and labeled by an unknown target $\ell$ in a hypothesis class
$\cH$. A positive-only learner sees only independent samples from $\cD_+$, the
distribution of $X\sim\cD$ given $\ell(X)=1$, but is evaluated on the full
distribution $\cD$. We study the following problem.

\vspace{-4pt}
\begin{problem}[Proper positive-only PAC learning]\label{prob:ppo}
Given a hypothesis class $\cH$, find a learner and a function $m_{\cH}$, called the learner's sample
complexity, that depends only on $\cH$, such that the following holds. For every
$\varepsilon,\delta\in(0,1)$, every distribution $\cD$, and
every target $\ell\in\cH$ with $\Pr_{X\sim\cD}[\ell(X)=1]>0$, the learner,
given $n\ge m_{\cH}(\varepsilon,\delta)$ independent samples
from $\cD_+$, returns a hypothesis $h\in\cH$ satisfying
$\Pr_{X\sim\cD}[h(X)\neq\ell(X)]\leq\varepsilon$, with probability at least
$1-\delta$ over the sample and any internal randomness of the learner.
\end{problem}

\vspace{-4pt}
\noindent Notice that the
learner never sees a negative example, and yet false positives also count
toward its error. If, for instance, the hypothesis class
contains the all-ones classifier, then this classifier is consistent with
every sample but errs on every point outside the target.

\cref{prob:ppo} asks for \emph{proper}
learning, which requires the output to belong to the same hypothesis class
as the target, so that it satisfies the same structural assumptions. For example, a proper learner for halfspaces must return
a halfspace. In ordinary binary PAC learning, properness comes for free, since empirical risk minimization is a deterministic proper learner
for every class of finite VC dimension, and these are exactly the learnable
classes. In positive-only learning, by contrast, properness makes the problem much harder. Without properness, the problem is well understood:
its characterization appears even as a textbook exercise
\citep[Exercise~3.7]{KearnsVazirani1994}, and every learnable class is learned by simply outputting the intersection of all hypotheses containing the sample. This output makes no false positives, but it usually lies outside the
class (for halfspaces, it is the convex hull of the sample). In fact, properness can change which classes are learnable: some classes are learnable by an improper learner but by no proper one
\citep{PositiveOnly}.

A characterization of proper learning remained open for decades after the work of
\citet{Natarajan1987}. Recently, \citet{PositiveOnly} settled the randomized
case by identifying necessary and sufficient combinatorial conditions for
proper learnability. They also showed that, unlike in PAC learning, randomness matters: some classes are
properly learnable by a randomized learner but by no deterministic one. With the improper and randomized cases settled, one basic case \mbox{remains open}:

\vspace{-9pt}
\begin{quote}
\centering
\emph{What
is the characterization of deterministic proper positive-only learning?}
\end{quote}
\endgroup

\subsection{Our Contributions}

We show that the answer can depend on the axioms of
mathematics: even for a class of VC dimension two, the existence of a
deterministic proper learner can be independent of the standard axioms of
set theory.

These axioms, called \emph{ZFC} (Zermelo--Fraenkel set theory
with the axiom of choice), provide the usual foundation for mathematics,
including learning theory
\citep{Jech}.
A statement is independent of ZFC if neither it nor its negation can be
proved from these axioms. The
best-known example is the continuum hypothesis, which we recall later in this section
\citep{Godel1940,Cohen1963}. For our class,
there are models of ZFC, that is, mathematical universes satisfying
all its axioms, in which a deterministic proper learner exists and
others in which it does not. Thus the standard tools
of learning theory cannot settle whether this class is deterministically learnable. We ask whether
any learning rule exists, even an uncomputable one.

Since we allow uncomputable learners, this notion of independence from ZFC is very
different from the computational undecidability of problems such as the
halting problem \citep{Turing1937}.
Arbitrary learning rules need not be measurable, so we consider
distributions with countable support, for which the guarantee in
\cref{prob:ppo} is well defined for every rule. Restrictions to
distributions with countable or finite support are standard in work on the
independence of learnability~\citep{FiniteCharacter,FishReyzin2020,HannekeYang2023}.
Accordingly, a class is \emph{deterministically learnable} if a learner whose
output is a function of the unordered sample satisfies this guarantee for
every such distribution.

\begin{informaltheorem}[Learnability Is Independent of ZFC; see \cref{thm:independence}]\label{thm:intro-independence}
Assuming ZFC is consistent, there is a class of VC dimension two whose
deterministic learnability is independent of ZFC.
\end{informaltheorem}

\noindent Here, determinism is necessary, as the same class is learnable by a randomized learner in ZFC.
\citet{FiniteCharacter,EMX2019} were the first to show that
learnability can be independent of ZFC. They studied a problem they called
\emph{estimating the maximum} (EMX), and showed that its learnability
depends on the cardinality of the continuum.
Since then, learnability in
several other problems has been shown to be independent of ZFC, including class-ratio learning
\citep{FishReyzin2020}, bandit learning \citep{HannekeYang2023}, and proper
multiclass learning \citep{AsilisEtAl2025}. These results build on EMX and its
monotone-compression machinery, which ties learnability to the cardinality
of the domain. In contrast, our proof
takes a different route and analyzes the learner directly using the Erd\H{o}s--Rado partition theorem \citep{ErdosRado}, an infinite analogue of
Ramsey's theorem.
We color finite subsets of the domain by the learner's behavior. On
a sufficiently large domain, the partition theorem yields a large subset on
which this behavior is uniform. Properness then restricts the learner's outputs enough to construct
a distribution on which it fails (see \cref{sec:technical-overview} for details).
This argument may also be useful for establishing nonlearnability beyond
this particular class.

Combinatorial characterizations, such as
finite VC dimension for PAC learning, express
learnability through properties of the hypothesis class. Ideally, these properties
should be as ``simple'' as possible. To capture one notion of simplicity,
\citet{FiniteCharacter,EMX2019} introduced \emph{finite character}: a
dimension has finite character if every lower bound on it is witnessed by
finitely many domain points and hypotheses. Most existing dimensions in
learning theory have this property. For instance, for the VC dimension, the
witness is a shattered set together with hypotheses realizing its labelings,
and the Natarajan, graph, and Littlestone dimensions behave similarly.

One might expect \cref{thm:intro-independence} to already rule out a
characterization by a dimension of finite character. However, passing to
another model of ZFC can change the domain or hypothesis class, which can
also change the dimension. Independence alone therefore does not rule out
such a characterization. Our next result compares two models with the same
domain and hypotheses, in which every dimension of finite character has the
same value \citep{FiniteCharacter}, and shows that, despite this, the
deterministic learnability of our class changes.

\begin{informaltheorem}[No Finite-Character Characterization; see \cref{thmt@@MainNoFinite}]\label{thm:intro-finite-character}
Assuming ZFC is consistent, there are two models of ZFC that contain
the same domain and hypotheses of the class in
\cref{thm:intro-independence}, yet this class is deterministically
learnable in one model but not the other. Consequently, no dimension of
finite character characterizes deterministic learnability.
\end{informaltheorem}

\noindent\textbf{Domain size and learnability.}
The partition argument behind
\cref{thm:intro-independence} needs a large domain. The required size grows with the
sample size, so the domain must be large enough for all sample sizes at
once. It is therefore natural to ask whether smaller domains avoid the obstruction:

\begin{quote}
\centering
\emph{Does bounding the domain size restore a characterization of learnability in ZFC?}
\end{quote}

\noindent We give such a characterization, with size bounds expressed
using two standard sequences of infinite cardinals, the
aleph and beth numbers. We review them briefly; \cref{sec:prelim-sets} gives more background.

\begin{itemize}[label=$\blacktriangleright$,leftmargin=1em,labelsep=.3em]
\item \textbf{Cardinals and the aleph numbers.}
The cardinal $\aleph_0$ is the size of the natural numbers. The next cardinal,
$\aleph_1$, is the smallest cardinal larger than $\aleph_0$, and
$\aleph_2,\aleph_3,\ldots$ are defined by repeatedly taking the next larger
cardinal. Their supremum, $\aleph_\omega$, is the smallest cardinal larger
than every $\aleph_n$ for finite $n$. The reals form an uncountable set, so their
cardinality is at least $\aleph_1$. The assertion that it is exactly
$\aleph_1$ is the \emph{continuum hypothesis} (CH).

\item \textbf{Power sets and the beth numbers.}
The reals have the same cardinality as the set of all subsets of the
natural numbers. This cardinality is denoted $\beth_1\coloneqq 2^{\aleph_0}$. Iterating power sets gives the beth numbers: $\beth_0=\aleph_0$ and $\beth_{n+1}=2^{\beth_n}$. Thus $\beth_1=|\R|$, and $\beth_2$ is the cardinality of the
set of all subsets of the reals. The supremum of this sequence is
$\beth_\omega$, and $\aleph_\omega\le\beth_\omega$. The \emph{generalized continuum hypothesis}
(GCH) says that,
for every infinite cardinal $\kappa$, $2^\kappa$ is the next larger
cardinal. It implies $\beth_n=\aleph_n$ for all finite $n$, and hence
$\beth_\omega=\aleph_\omega$. Both CH and GCH are independent of ZFC,
assuming its consistency \citep{Jech}.
\end{itemize}

\noindent Thus ZFC alone does not determine
whether the reals have size $\aleph_1$, nor whether
$\aleph_\omega=\beth_\omega$. \cref{thm:intro-domain} therefore states separately
what holds in ZFC and what follows under GCH.

\begin{informaltheorem}[Domain Size and Learnability; see \cref{thm:main-domain}]\label{thm:intro-domain}
The following hold.

\begin{enumerate}
\item In ZFC, on every domain of cardinality less than $\aleph_\omega$,
deterministic learnability is equivalent to the three combinatorial
conditions in \cref{def:intro-conditions} below.
\item In ZFC, every domain of cardinality at least $\beth_\omega$ supports
a class of VC dimension two that satisfies all three conditions in
\cref{def:intro-conditions} but is not deterministically learnable.
\item Under GCH, $\beth_\omega=\aleph_\omega$, so these bounds coincide:
the three conditions characterize deterministic learnability of every class
on a domain $\cX$ if and only if $|\cX|<\aleph_\omega$.
\end{enumerate}
\end{informaltheorem}

\noindent In ZFC alone, our bounds leave a possible gap between
$\aleph_\omega$ and $\beth_\omega$.
The domains in part~(ii) are much larger than those
common in learning theory: a set of cardinality $\beth_\omega$ is larger
than any set obtained from $\R$ by taking power sets finitely many times.
Even so, in our lower bound constructions each deterministic learner
fails on a distribution with finite support.

Next,
we describe the three conditions in part~(i) of \cref{thm:intro-domain}. We use the \emph{closure} of a sample, a standard learning-theoretic
notion.
We identify each hypothesis with the set of points
it labels positive. A set $S$ is \emph{realizable} if some hypothesis
contains it, and its \emph{version space}
$\cH_S=\{h\in\cH:S\subseteq h\}$ consists of all possible targets. Its \emph{closure} is the
intersection of all hypotheses in its version
space, $\CLOS_{\cH}(S)=\bigcap_{h\in\cH_S}h$, the points that $S$ forces to be
positive.

\begin{introdefinition}[Combinatorial conditions]\label{def:intro-conditions}
We consider the following conditions on $\cH$.

\begin{enumerate}
\item \emph{Finite VC dimension:} $\VCdim(\cH)<\infty$.
\item \emph{Uniform exterior separability:} For every $\eta>0$, there is
$M\in\Nat$ such that any finite nonempty realizable set $S$
admits $h_1,\ldots,h_M\in\cH_S$ with
$|\{j:y\in h_j\}|\le\eta M$ for every $y\notin\CLOS_{\cH}(S)$;
see \cref{eq:ues}.
\item \emph{Proper singleton closures:} For every realizable point $x$,
$\CLOS_{\cH}(\{x\})$ belongs to $\cH$.
\end{enumerate}
\end{introdefinition}
\noindent Informally, uniform exterior separability asks for a list of consistent hypotheses, of bounded length, in which each point outside the closure appears rarely. The first two conditions characterize learnability by randomized proper learners~\citep[Theorem~4]{PositiveOnly}, and deterministic learners also need the third~\citep[Proposition~18]{PositiveOnly}. Hence all three conditions are necessary on every domain (\cref{rem:intro-conditions} explains the role of each). \cref{thm:intro-domain} shows that the three conditions are also sufficient on every domain of
cardinality less than $\aleph_\omega$, but not on any domain of cardinality
at least $\beth_\omega$. On domains of cardinality $\aleph_\omega$, the gap between randomized
and deterministic learners can even depend on the axioms: the class in
\cref{thm:intro-independence} needs randomness in some models of ZFC but not
in others.

On domains of cardinality less than $\aleph_\omega$, \cref{thm:intro-domain} gives a
characterization of a different kind from those ruled out by
\cref{thm:intro-finite-character}. The restriction on the domain is essential: the class in \cref{thm:intro-independence} satisfies all three conditions in every model of ZFC, yet its deterministic learnability depends on the model. Each condition is a first-order statement whose quantifiers range over domain points and members
of $\cH$. In this sense they are combinatorial, and they meet the
requirement for a combinatorial characterization stated by
\citet{EMX2019}. However,
exterior separability and singleton closure do not appear to admit finite
evidence of failure, so they are not \emph{finitary} conditions in the
sense of \citet[Section~3.3]{LechnerBenDavid2024}.

\paragraph{A general obstruction.}
\cref{thm:intro-domain}~(ii) follows from a more general criterion for the
nonlearnability of a class.
\begin{proposition}[see \cref{thm:uniform-cardinal-condition}]\label[proposition]{thm:intro-obstruction}
Let $\cH$ be a concept class and $A$ be a realizable set with $|A|\ge\beth_\omega$ and $\CLOS_{\cH}(A)\notin\cH$. If, for some
cardinal $\kappa<\beth_\omega$, at most $\kappa$ hypotheses contain $A$ and
every other hypothesis meets $A$ in at most $\kappa$ points, then $\cH$ is
not deterministically learnable. 
\end{proposition}

\subsection{Technical Overview}\label{sec:technical-overview}

In this section, we highlight the key technical ideas behind our proofs.
After explaining why choosing a proper hypothesis without
randomness is difficult, we use a single class to illustrate both sides of our results: on large domains no deterministic
learner succeeds, while on smaller domains we construct one. We then extend these arguments
to general classes, and finally explain how the two sides combine to give
independence from ZFC.

\paragraph{An obstruction to proper positive-only learning.}
We begin with the difficulty that underlies
all our results. A positive-only learner is evaluated on the full
distribution, yet its sample $S$ never reveals whether an unobserved point is
negative. The only points it can safely label positive are those in the
closure $\CLOS_{\cH}(S)$, which every target consistent with $S$ labels
positive. Whenever this closure does not belong to $\cH$, a proper output cannot both
contain the sample and avoid all points outside the closure.
The learner must either output a hypothesis $h$ with
$h\setminus\CLOS_{\cH}(S)\ne\varnothing$, risking false-positive
error, or omit some observed positives, risking false-negative
error. Under UES, a randomized learner can remain consistent
with the sample while controlling the first risk: choosing
uniformly from the list $h_1,\ldots,h_M\in\cH_S$ in
\cref{def:intro-conditions} ensures
that every point outside the closure is included with
probability at most $\eta$.
Since the closure is contained in the target, every negative
point lies outside it, so averaging this
pointwise guarantee over the negative region bounds the expected
false-positive error. Consistency and finite VC dimension control the
false-negative error. \citet[Theorem~4]{PositiveOnly} show that UES and finite VC dimension characterize randomized proper learnability.

A deterministic learner cannot make an independent random
choice from this list. One strategy, used in
\citet[Theorem~7]{PositiveOnly} and in our positive results
discussed below, is to let the choice of hypothesis vary
across possible samples, using the randomness already present
in the data. The learner sees an unordered multiset, so its choice can depend only on the observed points and their multiplicities. This source of variation disappears when
$\cD_+$ is concentrated at a single point $x$: for each
sample size $n$, the learner always observes $n$ copies
of $x$. In this case, proper singleton closure is necessary
to avoid both errors \citep[Proposition~18]{PositiveOnly}.

\paragraph{A central construction.}
Proper singleton closure
handles samples concentrated at one point, but it does not resolve the
difficulty in general: the following class satisfies all three conditions,
yet whether it is deterministically learnable depends on the cardinality of
its domain.
The construction extends the example in \citet[Theorem~7]{PositiveOnly}. Let $A$ be an infinite
set, and let $B=\{b_1,b_2,\ldots\}$ be a countably infinite
set disjoint from $A$. Define
\begin{equation}\label{eq:example-class}
 \cX_A=A\sqcup B,\qquad
 h_j=A\cup\{b_j\}\quad(j\ge1),\qquad
 \mathcal M_A=
 \{\{x\}:x\in\cX_A\}\cup\{h_j:j\ge1\}.
\end{equation}
Each hypothesis $h_j$ consists of $A$ and one additional
\emph{marker} $b_j$, and we call $\mathcal M_A$ the \emph{marker class}. The class has VC dimension
two, and it satisfies UES and proper singleton closure.
However, any sample supported on $A$ with at least two
distinct points has closure $A$, which is not a hypothesis
in $\mathcal M_A$.
A proper learner must therefore either output a
singleton, discarding all but one observed point, or
output some $h_j$, which includes a
marker $b_j$ whose label the sample does not reveal.

\paragraph{Nonlearnability results.}
We first show that $\mathcal M_A$
is not deterministically learnable when $|A|\ge\beth_\omega$, which proves \cref{thm:intro-domain}~(ii), and then generalize the argument to \cref{thm:intro-obstruction}.

Consider $\mathcal M_A$ and samples supported on $A$.
If the positive distribution is spread
over many points, a singleton incurs large false-negative
error, while each $h_j$ risks a false positive at $b_j$.
To spread this risk over the markers, a deterministic learner must
therefore vary $j$ across samples. As there are only countably many $h_j$, the partition theorem rules this out when $A$ is large enough.

Fix a sample size $n$ and color
each $n$-element subset of $A$ by $0$ if the learner outputs
a singleton on it, and by $j$ if it outputs $h_j$. When
$|A|\ge\beth_\omega$, the Erd\H{o}s--Rado theorem gives an infinite subset $F\subseteq A$ on which all $n$-element subsets
have the same color. Such a set is called \emph{homogeneous}.
The theorem extends Ramsey's theorem to infinitely many colors, at
the price of a much larger ground set. The bound $|A|\ge\beth_\omega$ lets us apply it for every
fixed finite $n$. The set $F$ may depend on $n$ and on the learner. We take the positive distribution to be
uniform on a sufficiently large finite subset of $F$, so that
the sample has no repeated observations with high probability.
If the common color is $0$, the learner returns a singleton
and misses most of the positive mass. Otherwise, it always
returns the same $h_j$ on such samples. We can then choose
a target $h_{j+1}$ and place substantial negative mass on
$b_j$, forcing large false-positive error.
Thus, for every proposed sample size, a finitely supported
distribution defeats the learner (\cref{thm:marker-large}).

The proof of
\cref{thm:intro-obstruction} extends this argument to classes
whose hypotheses need not have the simple form above. Let $A$ and $\kappa$ be
as in that proposition, so that $|\cH_A|\le\kappa$ and $|h\cap A|\le\kappa$ for every
$h\notin\cH_A$.
Unlike the singleton outputs in $\mathcal M_A$, hypotheses
outside $\cH_A$ can now contain many points of $A$.
We therefore record the output's prediction at an omitted point, in addition to its type.

Again fix a sample size $n$, together with a well-order of $A$. We color
each $(n+1)$-element subset by the learner's behavior after deleting each
of its points, listing the deletions in this order. For each
deletion, we record which hypothesis in $\cH_A$ is returned,
or $*$ if the output lies outside $\cH_A$, together with
whether that output includes the deleted point. There are
at most $\kappa$ possible colors. The cardinal form of the
Erd\H{o}s--Rado theorem (\cref{thm:erdos-rado}) then
gives a sufficiently large homogeneous set $F\subseteq A$.

Homogeneity makes the
prediction at an unobserved point of $F$ depend only on its position relative to the
sample points. Choose a sample from $F$ with more than $\kappa$ points of $F$
in each gap before, between, and after its points. If the learner's output on this sample lies outside
$\cH_A$, its predictions are constant on each gap, so a
positive prediction on any gap would put more than $\kappa$ points of $A$
into a hypothesis outside $\cH_A$, contradicting $|h\cap A|\le\kappa$. All these predictions must therefore be negative.
Comparing the deletions also forces the output type to agree across all
$n$-element samples from $F$. Consequently, either the learner returns one fixed hypothesis in
$\cH_A$ on every such sample, or its output
contains no point of $F$ outside the sample. In the first
case, the fixed hypothesis contains a point outside
$\CLOS_{\cH}(A)$, which can be assigned substantial negative
mass. In the second, spreading the positive mass over many
points forces large false-negative error. Thus the argument
reduces to the same two types of failure as for
$\mathcal M_A$, again using only finitely supported
distributions (\cref{thm:uniform-cardinal-condition}).

\paragraph{Controlling false negatives.} The class $\mathcal M_A$
is deterministically learnable whenever $|A|\le|\R|$
(\cref{thm:marker-small}). This gives the learnable side of the
independence result. Every
class satisfying the three combinatorial conditions is
deterministically learnable when $|\cX|<\aleph_\omega$,
as asserted in \cref{thm:intro-domain}(i). Both proofs follow
the same strategy, similar to the one in
\citet[Theorem~7]{PositiveOnly}. For sufficiently large $n$, put $t=n^{2/3}$. If a point
$x$ occurs at least $n-t$ times, return
$\CLOS_{\cH}(\{x\})$; otherwise, return a hypothesis
consistent with the positive sample. The first output is proper by singleton closure and is contained in
the target, so it makes no false positives. It omits at most $t$
observations, while the consistent output omits none. Thus the empirical
false-negative error is at most $t/n=n^{-1/3}$, and VC uniform
convergence bounds the population false-negative error.
The main task is therefore to choose the consistent outputs
so that their false-positive error is also small.

\paragraph{Learning the marker class.}
After handling highly repeated points, an observed marker $b_j$ identifies
the target as $h_j$. The remaining samples are supported on $A$. We assign
each such sample to some $h_j$ so that
the probability of selecting any particular $j$ is small.

Fix an injection $\iota:A\to\R$. A \emph{rational cut}
is a rational threshold $r$ separating the observations
with $\iota(x)<r$ from those with $\iota(x)\ge r$.
This is where the bound $|A|\le|\R|$ is used: through $\iota$, the
countably many rational cuts can separate the points of every sample.
Choose the first rational cut, in a fixed enumeration, with at least
$t/2$ observations on each side, counting repetitions.
No point occurs $n-t$ times or more, so such a cut exists.
We encode both the cut's index and the number of observations
below it into the output index $j$. Fixing $j$ therefore
fixes a cut and a count. We
need not condition on which cut was selected: the event that this branch outputs $h_j$
is contained in the event that this fixed cut has this fixed count. The count is balanced, so a binomial bound gives probability $O(t^{-1/2})$.
Since $h_j$ adds only the point $b_j$ to
$A$, this controls the false-positive error.

\paragraph{Learning general classes on smaller domains.}
For general classes, we use UES to assign consistent hypotheses within finite groups of samples, which
the domain-size bound lets us form. The finite selection lemma (\cref{lem:selection}) provides
the required assignment: for any finite list of realizable samples
$S_1,\ldots,S_K$, we can simultaneously choose
$h_i\in\cH_{S_i}$ so that every point $y$ belongs to
$h_i\setminus\CLOS_{\cH}(S_i)$ for at most
$r=d^\star+1$ indices, where $d^\star$ is the VC dimension of the dual
class (\cref{sec:extended-prelim}). We call such samples \emph{bad
for $y$}. The bound holds simultaneously for every $y$ and is independent of
the length of the list. For a negative point $y$, the closure of every
positive sample excludes $y$, so these are exactly the samples on which
the selected output makes a false positive at $y$.
The proof is probabilistic:
choosing each $h_i$ independently from a UES list makes excessive overlap at
any fixed point unlikely. The bounded list sizes and finite dual VC
dimension limit the number of distinct overlap events, so a union bound at a
small enough UES scale yields a suitable tuple. We fix one such tuple before
observing the sample.

The assumption $|\cX|<\aleph_\omega$ supplies a finite reconstruction
map~\citep{FiniteCharacter}. Each sample can be represented by a message
consisting of at most $k$ observed points, for a fixed finite $k$, from
which a finite set containing its support can be reconstructed
(\cref{lem:reconstruction}). For example, on a countable domain, fix an
enumeration: the observed point with the largest index identifies a finite initial segment
containing the whole sample. Such maps exist exactly on domains of
cardinality less than $\aleph_\omega$ \citep{FiniteCharacter}, which is why
our characterization is stated for these domains. In general, we group samples of size $n$ by their
chosen message. Each group is finite because its samples have fixed size
and lie in one finite reconstructed set. Applying \cref{lem:selection}
within each group fixes a consistent output for every sample, with at most
$r$ bad samples for each point $y$ in each group.

To turn this into a probability bound, consider any finite set $U$ of
$m$ points. Samples supported on $U$ can use at most
$\Pi_k(m)=\sum_{j=0}^{\min\{k,m\}}\binom mj$ messages, so at most
$r\Pi_k(m)$ of them are bad for a fixed $y$. This count grows polynomially
in $m$, independently of the sample size $n$. The pooling lemma
(\cref{lem:pool}) converts this count into a probability bound that tends to zero, once samples with a highly repeated point are excluded. Its proof pools $2n$ observations and bounds the
chance that a random half of them forms a bad sample. This pooling is only an analysis device: the learner uses no additional samples or random bits. Averaging the resulting pointwise bound over the
negative region bounds the expected false-positive error. Together with the
earlier false-negative bound and Markov's inequality, this proves learnability (\cref{thm:domain}~(i)).

\paragraph{Independence from ZFC and finite character.}
We apply the positive and negative results to the same class in two
models of ZFC. Fix $A_\star=\aleph_\omega\times\{0\}$ and a disjoint
countable set of markers. Assuming ZFC is consistent, start in a model
satisfying GCH. In this model, $|A_\star|=\aleph_\omega=\beth_\omega$, so
$\mathcal M_{A_\star}$ is not deterministically learnable.

A standard forcing construction extends this model by adding new sets,
including new real numbers, while preserving cardinals and the elements
of the old domain and hypotheses. In the extension,
$|\R|>\aleph_\omega$, so our rational-cut
construction gives a learner for the
same class. The class stays the same, and only the set of available learning rules grows: the extension adds new functions,
such as an injection of $A_\star$ into $\R$, but no new points or
hypotheses. This establishes
\cref{thm:intro-independence}; see \cref{sec:independence-proof} for the
set-theoretic details.

Finally, a property whose quantifiers range only over the unchanged domain
points and hypotheses has the same truth value in both models. Such a
property cannot characterize deterministic learnability, which differs between the two models.
The models also have the same natural numbers, and each lower bound on a dimension of finite character has the same truth value in both. Hence such a dimension is finite in both models or in neither, proving \cref{thm:intro-finite-character}.

\section{Preliminaries}\label{sec:prelim}

In this section, we introduce notation and key preliminaries
(also see \cref{sec:extended-prelim}).

\medskip\noindent\phantomsection\textbf{Notation.}\label{sec:notation}
We write $\Nat=\{1,2,\ldots\}$ and $\omega=\Nzero=\{0,1,2,\ldots\}$. The notation $[A]^n$ denotes the family of subsets of $A$ of size $n$.
The indicator $\1\{E\}$ is one when $E$ holds and zero otherwise.
For sets $A,B$, their symmetric difference is
$A\triangle B=(A\setminus B)\cup(B\setminus A)$.
Write $\cX^{[*]}$ for the collection of finite unordered multisets
over $\cX$. For a multiset $S$, let $|S|$ denote its size, including
repetitions, and let $\supp(S)=\{x: x \in S \}$ denote its set of distinct
points. 

\subsection{\texorpdfstring{Positive-only learning}{Positive-only learning}}\label{sec:prelim-learning}

\medskip\noindent\textbf{The learning model.}
We now make the model of \cref{prob:ppo} precise.
Let $(\cX,\Sigma)$ be a measurable space whose $\sigma$-algebra contains
all singletons, and let
$\varnothing\ne\cH\subseteq\Sigma$ be a concept class; we identify each
hypothesis with its positive region.
Let $\cP$ be a family of probability distributions on this space.
\cref{prob:ppo} corresponds to the family of all distributions, and our results concern countably supported distributions (\cref{asm:countable-support} below).
For a target $\ell\in\cH$ and a distribution $\cD\in\cP$ with
$\cD(\ell)>0$, the learner receives samples from
$\cD_+=\cD(\,\cdot\mid\ell)$, and the error of $h\in\cH$ is
$\err_{\cD}(h,\ell)=\cD(h\triangle\ell)=\Pr_{X\sim\cD}[h(X)\neq\ell(X)]$.

\begin{assumption}\label{asm:countable-support}
$\cP$ is the family of countably supported distributions on $\cX$, that is, distributions
$\cD$ for which some countable set $C\subseteq\cX$ has $\cD(C)=1$.
\end{assumption}

\begin{remark}[Choice of distributions]\label{rem:distributions}
\normalfont
Our results quantify over all learning rules, including nonmeasurable ones, and for countably supported distributions the guarantee in \cref{prob:ppo} is well defined for every rule. Similar restrictions are standard~\citep{MoranYehudayoff}, especially in work on the
independence of learnability from ZFC~\citep{FiniteCharacter}, \citep[Corollary~8]{FishReyzin2020},
\citep[Appendix~C.1]{HannekeYang2023}. This restriction does not affect our lower bounds, which use only finitely supported distributions and hence hold for every family of distributions that contains them.
The characterization in \cref{thm:main-domain}(i) also extends to all distributions if $\cH$ satisfies the measurability conditions under which VC uniform convergence holds \citep{KrappWirth} and the learner constructed in its proof has jointly measurable predictions, that is, the map $(x_1,\ldots,x_n,y)\mapsto\1\{y\in\mathcal A(\{x_1,\ldots,x_n\})\}$ is measurable.
\end{remark}

\medskip\noindent\textbf{Learners.}
The learner observes an unordered sample: a multiset of $n$
independent draws from $\cD_+$, written $S\sim\cD_+^n$.
A deterministic proper learner is a function
$\mathcal A:\cX^{[*]}\to\cH$. It is said to be a PAC learner if there is a sample bound
$m_{\cH}:(0,1)^2\to\Nat$ such that, for every $\ell\in\cH$,
$\cD\in\cP$ with $\cD(\ell)>0$, $\varepsilon,\delta\in(0,1)$, and
$n\ge m_{\cH}(\varepsilon,\delta)$,
\begin{equation}\label{eq:pac}
 \Pr_{S\sim\cD_+^n}
 [\err_{\cD}(\mathcal A(S),\ell)>\varepsilon]\le\delta.
\end{equation}
Properness requires only
$\mathcal A(S)\in\cH$; the output need not contain the sample.
A \emph{randomized} proper learner also receives a uniformly random
seed $r\in\{0,1\}^{\Nat}$, independent of the sample, so that
$\mathcal A:\cX^{[*]}\times\{0,1\}^{\Nat}\to\cH$; the probability
in \cref{eq:pac} is then over both $S$ and $r$. We call $\cH$
\emph{deterministically learnable} if it has a deterministic proper PAC
learner, and \emph{learnable by a randomized learner} if it has a
randomized one. The restriction to multisets matters only for
deterministic learners: given the multiset, the order of an i.i.d.\ sample
without repeated points is a uniformly random permutation, which a deterministic learner could otherwise use as a source of random bits.

\medskip\noindent\textbf{Version spaces and closures.}
For a set $A\subseteq\cX$, its \emph{version space} is
$\cH_A=\{h\in\cH:A\subseteq h\}$.
For a multiset $S$, set $\cH_S=\cH_{\supp(S)}$.
A set or sample is \emph{realizable} if its version space is nonempty.
For a nonempty version space $V$, let $L(V)=\bigcap_{h\in V}h$, and define
\[
 \CLOS_{\cH}(A)=L(\cH_A)\qquad\text{and}\qquad
 \CLOS_{\cH}(S)=L(\cH_S)
\]
whenever the corresponding version space is nonempty.
Since a sample from $\cD_+$ lies in the target almost surely,
$\ell\in\cH_S$ and hence $\CLOS_{\cH}(S)\subseteq\ell$. Conversely, each $y\notin\CLOS_{\cH}(S)$ is
excluded by some $h\in\cH_S$, so the sample cannot rule out that $y$ is
negative. The closure is thus exactly the set of points that the
sample forces to be positive. An improper learner can output the closure
\citep[Exercise~3.7]{KearnsVazirani1994}, but the closure need not belong
to $\cH$. For example, if $\cH$ consists of the closed halfspaces in
$\R^d$, then the closure of a finite nonempty set is its convex hull, which
is not a halfspace.

\medskip\noindent\textbf{The combinatorial conditions.}
A class $\cH$ has \emph{uniform exterior separability} (UES)
\citep{PositiveOnly} if, for every $\eta>0$, there is $M(\eta)\in\Nat$ such that every finite nonempty realizable $S$ admits
$h_{S,1},\ldots,h_{S,M(\eta)}\in\cH_S$ satisfying
\begin{equation}\label{eq:ues}
 \sup_{y\notin\CLOS_{\cH}(S)}
 \frac1{M(\eta)}\sum_{j=1}^{M(\eta)}\1\{y\in h_{S,j}\}\le\eta.
\end{equation}
The hypotheses in this list may repeat. We say that $\cH$ has
\emph{proper singleton closures} if $\CLOS_{\cH}(\{x\})\in\cH$
for every realizable $x$. Both conditions relax the
requirement that every closure belong to $\cH$. Under that requirement, UES
holds with $M(\eta)=1$ and $h_{S,1}=\CLOS_{\cH}(S)$. In general, it asks
only for a bounded list of hypotheses in which each point outside the
closure appears rarely. The condition of proper singleton closures keeps the requirement,
but only for single points. Together with finite VC
dimension, these are the three conditions of \cref{def:intro-conditions} (also see \cref{rem:intro-conditions}).

\begin{remark}[Discussion of \cref{def:intro-conditions}]\label{rem:intro-conditions}
\normalfont
Finite VC dimension
characterizes ordinary binary PAC learning \citep{KearnsVazirani1994}, and
positive-only learning provides less information, so it remains necessary.
\cref{sec:technical-overview} explains the roles of the other two
conditions: uniform exterior separability lets a randomized learner control
false positives, and proper singleton closure handles a positive
distribution concentrated at one point, where the sample provides no
randomness. Without proper singleton closures,
every deterministic proper learner fails on a distribution supported on at
most two points \citep[Proposition~18]{PositiveOnly}. A randomized learner can instead ``hedge''
between hypotheses, which is why this condition is absent from the randomized characterization of \citet{PositiveOnly}.
\end{remark}

\subsection{\texorpdfstring{Set-theoretic background}{Set-theoretic background}}\label{sec:prelim-sets}

Our results depend on the sizes of infinite sets, and some depend on
axioms beyond ZFC. We review relevant background, assuming no familiarity with set theory. For a comprehensive treatment, we refer the reader to \citet{Jech}. Additional background on ZFC and independence appears in
\cref{sec:extended-prelim}.

\medskip\noindent\textbf{Cardinals.}
The \emph{cardinality} $|A|$ of a set $A$ measures its size: $|A|=|B|$
if there is a bijection between $A$ and $B$, and $|A|\le|B|$ if there is an
injection from $A$ into $B$. Under the axiom of choice, any two
cardinalities are comparable, and every nonempty set of cardinals has a
least element. For a cardinal $\kappa$, its \emph{successor} $\kappa^+$ is
the least cardinal strictly larger than $\kappa$.
We use the standard notation:
\[
 \aleph_0=|\Nat|,\qquad \aleph_{j+1}=\aleph_j^+,\qquad
 \aleph_\omega=\sup\nolimits_{j<\omega}\aleph_j.
\]
Thus an infinite set has
cardinality less than $\aleph_\omega$ if and only if its cardinality is
$\aleph_j$ for some finite $j$. Unlike each $\aleph_j$ with $j\ge1$, the cardinal
$\aleph_\omega$ is a countable union of sets of smaller cardinality. It is
the least uncountable cardinal with this property.

\medskip\noindent\textbf{Power sets and the beth numbers.}
For a set of cardinality $\kappa$, $2^\kappa$ is the cardinality of
its power set. Cantor's theorem says that $2^\kappa>\kappa$ for every cardinal $\kappa$. The reals
have the cardinality of the power set of $\Nat$. Define
\begin{equation}\label{eq:beth}
 \beth_0=\aleph_0,\qquad \beth_{j+1}=2^{\beth_j},\qquad
 \beth_\omega=\sup\nolimits_{j<\omega}\beth_j,\qquad
 \cc=2^{\aleph_0}=|\R|=\beth_1.
\end{equation}
Thus $\beth_2$ is the cardinality of the power set of $\R$, and a set of
cardinality $\beth_\omega$ is larger than any set obtained from $\R$ by
taking power sets finitely many times. Since $\beth_{j+1}>\beth_j$,
induction gives $\aleph_j\le\beth_j$ for every $j$, and hence
$\aleph_\omega\le\beth_\omega$.

\medskip\noindent\textbf{The generalized continuum hypothesis (GCH).}
The \emph{continuum hypothesis} (CH) asserts that $\cc=\aleph_1$, that
is, every uncountable set of reals has the cardinality of $\R$.
The generalized continuum hypothesis (GCH) is the assertion that
$2^\kappa=\kappa^+$ for every infinite cardinal $\kappa$. G\"odel showed that GCH, and hence CH, is
consistent with ZFC, and Cohen showed that the negation of CH is consistent
with ZFC \citep{Godel1940,Cohen1963}. Thus both CH and GCH are independent
of ZFC. Already the identities $2^{\aleph_j}=\aleph_{j+1}$ for finite $j$
imply $\beth_j=\aleph_j$ and $\beth_\omega=\aleph_\omega$.
So under GCH, the two thresholds in
\cref{thm:intro-domain} coincide. Without GCH, they can differ: if $\cc>\aleph_\omega$, then
$\aleph_\omega<\cc=\beth_1<\beth_\omega$.
It is consistent with ZFC that $\cc=\aleph_j$ for any finite $j\ge1$,
and also that $\cc>\aleph_\omega$ \citep{Jech}.
So ZFC does not decide whether $\R$, or any domain of
cardinality $\cc$, has cardinality less than $\aleph_\omega$. One value is ruled
out: by K\"onig's theorem, $\cc$ is
not a countable union of sets of smaller cardinality \citep{Jech}, so
$\cc\ne\aleph_\omega$.

\medskip\noindent\textbf{Well-orders and ordinals.}
A \emph{well-order} is a linear order in which every nonempty subset has a
least element. The axiom of choice implies that every set admits one. We
identify each infinite cardinal with its \emph{initial ordinal}, the least
ordinal of that cardinality. \cref{sec:extended-prelim} reviews these
notions.

\medskip\noindent\textbf{The Erd\H{o}s--Rado partition theorem.}
A \emph{coloring} of $[A]^n$ is a map from $[A]^n$ to a set of `colors'.
A set $F\subseteq A$ is \emph{homogeneous} for a coloring $\chi$ if $\chi$
is constant on $[F]^n$. The following extension of Ramsey's theorem,
discussed in \cref{sec:technical-overview}, uses iterated exponentials: for an infinite cardinal $\lambda$, define
$\exp_0(\lambda)=\lambda$ and $\exp_{j+1}(\lambda)=2^{\exp_j(\lambda)}$, so
that $\exp_j(\aleph_0)=\beth_j$.
\begin{theorem}[Erd\H{o}s--Rado {\citep{ErdosRado,Jech}}]\label{thm:erdos-rado}
Let $\lambda$ be an infinite cardinal and $n\ge0$. If
$|A|\ge\exp_n(\lambda)^+$, then every coloring of $[A]^{n+1}$ with at most
$\lambda$ colors has a homogeneous set of cardinality $\lambda^+$.
\end{theorem}
\noindent Taking $\lambda=\aleph_0$ gives the following special case,
which suffices for part~(ii) of \cref{thm:intro-domain}.
\begin{corollary}[Infinite homogeneous sets]\label{lem:erdos-rado}
For every finite  $n\ge1$, every set $A$ with $|A|\ge\beth_{n-1}^+$, and
every $\chi:[A]^n\to\Nzero$, there is an infinite $F\subseteq A$
for which $\chi$ is constant on $[F]^n$. In particular,
$|A|\ge\beth_\omega$ suffices for every finite $n$.
\end{corollary}
\noindent The last assertion uses $\beth_{n-1}^+\le\beth_n<\beth_\omega$.
The size requirement in \cref{lem:erdos-rado} grows with $n$, and
some growth is necessary. For $n=2$, color each pair of distinct infinite
binary sequences by the first position where they differ. This coloring
uses countably many colors, yet no three sequences have all three pairs of
the same color. So a set of cardinality $\cc=\beth_1$ need not contain a
homogeneous set of size three, while \cref{lem:erdos-rado} applies to
sets of cardinality $\beth_1^+$.

\section{Formal Statements of Our Results}\label{sec:formal}

In this section, we present formal statements of our results.

Throughout, $\cP$ is the family of countably supported
distributions (\cref{asm:countable-support}). As in
\cref{sec:prelim-learning}, a class is \emph{deterministically learnable}
if some deterministic proper learner achieves the guarantee of
\cref{prob:ppo} for every distribution in $\cP$, and
\emph{learnable by a randomized learner} if some randomized proper learner
does.

We begin with the formal version of \cref{thm:intro-independence}, whose proof appears in \cref{sec:independence-proof}.
It concerns a specific instance of the class $\mathcal M_A$
in \cref{eq:example-class}, whose domain has cardinality $\aleph_\omega$.

\begin{restatable}[Learnability Is Independent of ZFC]{theorem}{MainIndependance}
\label{thm:independence}
Let $\cH_\star=\mathcal M_{A_\star}$ be the class
in \cref{eq:example-class} with $A_\star=\aleph_\omega\times\{0\}$ and
$B=\Nat\times\{1\}$. Then $\cH_\star$ has VC dimension two, and the
following hold.

\begin{enumerate}[label=(\roman*)]
    \item In ZFC, $\cH_\star$ is learnable by a randomized learner.
    \item Assuming ZFC is consistent, deterministic learnability of $\cH_\star$
    is independent of ZFC.
\end{enumerate}
\end{restatable}

\noindent Thus ZFC cannot decide whether this class, despite having
VC dimension two, is deterministically learnable, even though ZFC proves that
a randomized learner exists. In particular, the class needs randomness in
some models of ZFC but not in others, so whether randomness is necessary can
itself depend on the axioms.

Next, we formalize
\cref{thm:intro-finite-character}. We define finite
character, following \citet{FiniteCharacter,EMX2019}. Combinatorial
characterizations, such as finite VC dimension for PAC learning, are useful
because they are simple, and finite character captures one notion of
simplicity. Informally, a dimension has finite character if every lower
bound on it is witnessed by finitely many domain points and hypotheses. For
example, $\VCdim(\cH)\ge s$ holds if and only if some $s$ points and $2^s$
hypotheses in $\cH$ realize all labelings of these points. Formally, a
\emph{property} is a formula
$P(\cX,\cH)$ in the first-order language of set theory, whose only
relations are equality and membership, with free variables $\cX$ and
$\cH$. We think of $\cX$ as a domain and of $\cH$ as a class over it. A
property has \emph{finite character} if ZFC proves that it is equivalent to
a \emph{bounded formula}, that is, a formula with free variables $\cX$ and $\cH$ in which every quantifier has the form $\exists x\in\cX$,
$\forall x\in\cX$, $\exists h\in\cH$, or $\forall h\in\cH$. \\
A
\emph{dimension} $\mathrm D$ assigns a value
$\mathrm D(\cX,\cH)\in\Nzero\cup\{\infty\}$ to each class $\cH$ over a
domain $\cX$; it has finite character if there is a computable sequence $(\varphi_s)_{s\in\Nat}$ of
    bounded formulas such that ZFC proves the following statement: for every $s\in\Nat$ and
    all $\cX,\cH$, we have $\mathrm D(\cX,\cH)\ge s$ if and only if $\varphi_s$ holds for
    $(\cX,\cH)$. Such
properties are natural because their truth depends only on the domain and
the hypotheses themselves: a property of finite character has the same truth
value in any two models of ZFC that contain the same domain and the same
hypotheses \citep{FiniteCharacter}. A property $P$
\emph{characterizes} deterministic learnability if ZFC proves that, for
every domain $\cX$ and class $\cH$ over $\cX$, $P(\cX,\cH)$ holds if and
only if $\cH$ is deterministically learnable. A dimension $\mathrm D$
characterizes it if the property $\mathrm D(\cX,\cH)<\infty$ does.
We can now state the formal version of
\cref{thm:intro-finite-character}, whose proof appears in \cref{sec:finite-character-proof}.

\begin{restatable}[No Finite-Character Characterization]{theorem}{MainNoFinite}
\label{thm:no-finite-character}
Assume that ZFC is consistent. Then no property or dimension of finite
character characterizes deterministic learnability.
\end{restatable}

\noindent\cref{thmt@@MainNoFinite} shows that,
unlike PAC learning, deterministic proper positive-only learning admits no
characterization of the kind that VC dimension provides. Its proof uses the
two models from the proof of \cref{thm:independence}, which contain the same
domain and class but disagree on whether the class is deterministically
learnable. No property of finite character can distinguish these models.
Neither can a dimension of finite character: each property
$\mathrm D(\cX,\cH)\ge s$ has the same truth value in both models, and the
two models have the same natural numbers, so the dimension is finite in one
model if and only if it is finite in the other.

Our third result, the formal version of
\cref{thm:intro-domain}, concerns the size of the domain.
The proofs of parts~(i) and~(ii) appear in
\cref{sec:pos-proof} and \cref{sec:independence-proof}, respectively.

\begin{restatable}[Domain Size and Learnability]{theorem}{MainDomain}
\label{thm:main-domain}\label{thm:domain}
The following hold.

\begin{enumerate}[label=(\roman*)]
\item In ZFC, for every domain $\cX$ with
$|\cX|<\aleph_\omega$, every class
$\cH\subseteq2^{\cX}$ is deterministically learnable
if and only if it satisfies the three conditions in
\cref{def:intro-conditions}. If these conditions hold and
$|\cX|\le\aleph_{k-1}$ for some $k\ge1$, then
$O(2^{3d}(k+1)^6/(\varepsilon\delta)^3)$ samples suffice, where
$d=\VCdim(\cH)$.
\item In ZFC, for every set $A$ with $|A|\ge\beth_\omega$, the class
$\mathcal M_A$ in \cref{eq:example-class} has VC dimension two and
satisfies the three conditions in \cref{def:intro-conditions},
but is not deterministically learnable. In particular, every domain of
cardinality at least $\beth_\omega$ supports such a class.
\item Under GCH, $\beth_\omega=\aleph_\omega$. Hence, for every domain $\cX$, the
three conditions in \cref{def:intro-conditions} characterize
deterministic learnability of all classes over $\cX$ if and only if
$|\cX|<\aleph_\omega$.
\end{enumerate}
\end{restatable}

\noindent\cref{thm:main-domain} complements \cref{thm:independence},
whose domain has cardinality exactly $\aleph_\omega$. On every smaller
domain, including every countable one, ZFC proves a combinatorial
characterization of deterministic learnability, and the only condition
beyond the randomized characterization of \citet{PositiveOnly} is proper
singleton closures. On domains of cardinality at least $\beth_\omega$,
these conditions no longer suffice, and under GCH the two thresholds meet.

The nonlearnability in part~(ii) is an instance of the following
general criterion, the formal version of
\cref{thm:intro-obstruction}.
Its proof appears in \cref{sec:uniform-cardinal-proof}.

\begin{restatable}[Nonlearnability from Cardinality Conditions]{theorem}{MainCardCondition}
\label{thm:uniform-cardinal-condition}
Let $\cH$ be a class over a domain $\cX$. Suppose that
some realizable set $A\subseteq\cX$ satisfies
\vspace{1.5mm}

\begin{equation}\label{eq:uniform-cardinal-condition}
 |A|\ge\beth_\omega,\quad
 \CLOS_{\cH}(A)\notin\cH,\quad\text{and}\quad
 \max\{|\cH_A|,\sup\nolimits_{h\in\cH\setminus\cH_A}|h\cap A|\}<\beth_\omega\,.
\end{equation}
Then $\cH$ is not deterministically learnable.
For every deterministic proper learner
$\mathcal A$ and every sample size $n\ge1$, there are $\ell\in\cH$
and a finitely supported distribution $\cD$ with $\cD(\ell)>0$ such that
\vspace{1.5mm}
\begin{equation}\label{eq:uniform-cardinal-error}
 \Pr_{S\sim\cD_+^n}
 \left[\err_{\cD}(\mathcal A(S),\ell)\ge\nicefrac{1}{2}\right]\ge\nicefrac{3}{4}\,.
\end{equation}
\end{restatable}

\noindent\cref{thm:uniform-cardinal-condition} makes no assumption on
the VC dimension or on exterior separability. Its conditions involve only
cardinalities. The failure it gives is also strong: at every sample size,
the learner has error at least $1/2$ with probability at least $3/4$ on a
finitely supported distribution. We expect this criterion to be useful for
proving nonlearnability of other classes on large domains.

\section{Conclusion}

In this work, we study the question left open by the characterization
of randomized proper positive-only learning: which classes can be learned
without randomness? We show that the answer can
depend on the axioms of mathematics: assuming ZFC is consistent, there is a
class of VC dimension two whose deterministic learnability is independent of
ZFC, even though it is learnable by a randomized learner. Our results
also rule out a characterization by a dimension of finite character.
In ZFC, on domains of cardinality less than $\aleph_\omega$, we give a
characterization by three combinatorial conditions, and these
conditions do not suffice on any domain of cardinality at least
$\beth_\omega$. Under GCH, $\aleph_\omega=\beth_\omega$, so the threshold $\aleph_\omega$
on the domain size is sharp.

Our work leaves several questions open. In ZFC alone (without GCH), our bounds
leave a possible gap between $\aleph_\omega$ and $\beth_\omega$. More broadly, is
there a characterization of deterministic learnability on all domains? By
\cref{thm:intro-finite-character}, it cannot be a dimension of finite
character. What is the optimal sample
complexity of deterministic proper positive-only learning on domains of
cardinality less than $\aleph_\omega$? Can the Erd\H{o}s--Rado partition
argument establish independence from ZFC in other learning models?

\section*{AI Disclosure}

The authors developed most ideas in this paper. GPT-6-Pro suggested using the Erd\H{o}s--Rado partition theorem
to prove the nonlearnability side of \cref{thm:intro-independence}.

\printbibliography

\appendix

\clearpage

\section{Extended Preliminaries}\label{sec:extended-prelim}

In this section, we present additional preliminaries used in our proofs.

\smallskip\noindent\textbf{Additional notation.}
We write $[K]=\{1,\ldots,K\}$ for $K\in\Nat$.
We write $[A]^{\le k}$ and $[A]^{<\omega}$ for the families of
subsets of $A$ of size at most $k$ and of finite size, respectively.
For a multiset $S$, let $N_x(S)$ denote the multiplicity of $x$ in $S$.
For $x\in\cX$, $\delta_x$ is the point mass at $x$, and $\Unif(F)$ is the
uniform distribution on a finite set $F$.

\smallskip\noindent\textbf{VC dimension and the dual class.}
We write $d=\VCdim(\cH)$. The \emph{trace} of $\cH$ on a set $U$ is
$\{h\cap U:h\in\cH\}$. The dual class
$\cH^\star=\{e_x:x\in\cX\}$ has domain $\cH$, with
$e_x(h)=\1\{x\in h\}$. We write $d^\star=\VCdim(\cH^\star)$.
We also use
\[
 \Pi_k(m)=\sum_{j=0}^{\min\{k,m\}}\binom mj.
\]
Sauer's lemma bounds the size of the trace of $\cH$ on a finite set $U$ by
$\Pi_d(|U|)$~\citep{Sauer}, and Assouad's bound gives $d^\star<2^{d+1}$ when $d<\infty$
\citep{Assouad}.

\smallskip\noindent\textbf{ZFC and independence.}
ZFC denotes Zermelo--Fraenkel set theory with the axiom of choice,
the standard axioms of mathematics. Our learnability and nonlearnability results are theorems of ZFC, except
where we explicitly assume an additional hypothesis, such as GCH. A statement is \emph{independent} of ZFC if ZFC proves neither the
statement nor its negation. By G\"odel's completeness theorem, this holds
exactly when there are two \emph{models} of ZFC, that is, two mathematical
universes satisfying all its axioms, such that the statement is true in one
and false in the other. Independence results assume that ZFC is
consistent, since an inconsistent theory proves every statement. Our independence
results use Cohen's \emph{forcing}, the standard method for building models
\citep{Cohen1963,Jech}. It extends a given model by new sets, such as new
real numbers, while each old set keeps its elements. So a class and its
domain are the same in a model and in its extension, as
\cref{thm:intro-finite-character} requires. We use
extensions that are \emph{cardinal preserving}: every cardinal of the old
model remains a cardinal, so $\aleph_\omega$ (defined in \cref{sec:prelim-sets}) is the same in
both models. The domain of our class then keeps cardinality $\aleph_\omega$,
while the continuum can grow beyond it (\cref{sec:independence-proof}).

\smallskip\noindent\textbf{Well-orders and ordinals.}
A well-order is
a linear order in which every nonempty subset has a least element. For example,
the usual order on $\Nat$ is a well-order, but the usual orders on
$\mathbb Z$ and on $[0,1]$ are not. The axiom of choice 
implies that every set
admits a well-order. For uncountable sets such as $\R$, however, no
explicit well-order is known, and subsets defined in terms of such an ordering need not be
Lebesgue measurable. \emph{Ordinals} are canonical well-ordered sets:
every well-ordered set is order isomorphic to a unique ordinal, called its
\emph{order type}. Each ordinal is the set of smaller ordinals. The finite
ordinals are $0,1,2,\ldots$, and $\omega$ is the least infinite ordinal.
We
identify each infinite cardinal with its \emph{initial ordinal}, the least
ordinal of that cardinality, and write $\omega_j$ and $\omega_\omega$ for
the initial ordinals of $\aleph_j$ and $\aleph_\omega$. Thus a set $A$ of
cardinality $\kappa$ can be enumerated as $\{a_\alpha:\alpha<\kappa\}$, just
as a countable set can be listed as $a_0,a_1,\ldots$. In such an
enumeration, every initial segment $\{a_\beta:\beta<\alpha\}$ with
$\alpha<\kappa$ has cardinality less than $\kappa$. This property
drives the finite reconstruction in \cref{lem:reconstruction}.

\section{Related Work}\label{sec:related-work}

In this section, we present related models of learning from positive examples and further work on independence and characterizations of learnability.

\smallskip\noindent\textbf{Learning with one-sided error.}
\citet{Natarajan1987} initiated the study of learning from positive
examples in a model with one-sided error, in which the learner must never
label a negative point as positive. He characterized proper learnability in
this model, and \citet{Shvaytser1990} and \citet{Kivinen1995} characterized
improper learnability and bounded its sample complexity. Forbidding false
positives makes this model more restrictive than ours, in which the learner
may err in both directions and is evaluated by its total error.
\citeauthor{Natarajan1987} conjectured that his characterization extends to
proper learning in our two-sided model. \citet{PositiveOnly} refuted this
conjecture.

\smallskip\noindent\textbf{Learning from positive examples.}
In the two-sided model studied here, improper learnability is
characterized by finite VC dimension of the class of finite intersections of
hypotheses \citep{GerebGraus1989}. This characterization appears as an exercise in the textbook
of \citet[Exercise~3.7]{KearnsVazirani1994}. \citet{Hanneke2024Star}
determines the optimal sample complexity of improper learning in terms of
the $1$-centered star number. Both results concern improper learners, which
may output the closure of the sample even when it lies outside the class.
For proper learning, \citet{PositiveOnly} characterized learnability by
randomized learners, separated randomized from deterministic learners, and
studied consistent and stable learners. They left open the characterization
for deterministic learners, which is the question we study.

\smallskip\noindent\textbf{Distributional assumptions.}
Another line of work studies positive-only learning under
assumptions on the distribution. For Gaussian distributions,
\citet{KontonisTzamosZampetakis2019} and \citet{LeeMehrotraZampetakis2024}
give efficient algorithms for classes with bounded Gaussian surface area.
\citet{LeeMehrotraZampetakis2026} learn every class of finite VC dimension
when the unknown distribution satisfies a smoothness condition. Other works
assume that the underlying distribution is known, for example, uniform
\citep{DeDiakonikolasServedio2015,CanonneDeServedio2020}. These works obtain
efficient algorithms but rely on assumptions about the distribution. In
contrast, our results are distribution-free, and we ask which classes admit
a learner at all, without regard to computational efficiency.

\smallskip\noindent\textbf{Learning from positive and unlabeled examples.}
In positive-unlabeled (PU) learning, the learner also receives
unlabeled samples from the full distribution
\citep{Denis1998,LiuEtAl2002,ElkanNoto2008,BlanchardLeeScott2010,duPlessisNiuSugiyama2015,BekkerDavis2020,LeeMehrotraZampetakis2024,LeeMehrotraZampetakis2026}.
Unlabeled samples show where the distribution places mass, and hence where
false positives would be costly. \citet{MansouriBenDavid2025} give upper and
lower bounds on the numbers of positive and unlabeled samples needed for PU
learning. Their lower bounds show that, for some classes, no number of
positive examples can replace the unlabeled ones. Our learners receive no
unlabeled samples, so they can control false positives only through the
structure of the class.

\smallskip\noindent\textbf{Independence of learnability.}
In the EMX problem of \citet{FiniteCharacter,EMX2019}, discussed in
the introduction, the learner must output a member of a class whose
probability mass is close to the largest in the class. For the class of
finite subsets of $[0,1]$, they show that EMX learnability is equivalent to
the existence of a monotone compression scheme, and that both hold if and
only if $2^{\aleph_0}<\aleph_\omega$, which ZFC does not decide. Independence
results have since been established for class-ratio learning
\citep{FishReyzin2020}, bandit learning \citep{HannekeYang2023}, and proper
multiclass learning \citep{AsilisEtAl2025}, each by relating the problem to
EMX. Our independence proof instead analyzes the learner directly, using the Erd\H{o}s--Rado partition theorem
(\cref{sec:technical-overview}). We expect this approach to be useful for other learning problems, since it does not require relating them
to EMX. Our learnability results do use the
finite-reconstruction lemma of \citet{FiniteCharacter}, which underlies
their compression schemes and holds on every domain of cardinality less
than $\aleph_\omega$, the same cardinal that governs EMX learnability.

\smallskip\noindent\textbf{Limits of dimension-based characterizations.}
\citet{FiniteCharacter,EMX2019} also showed, by comparing two models
of ZFC, that no dimension of finite character characterizes EMX
learnability. \cref{thm:intro-finite-character} is the analogous result for
deterministic proper positive-only learning. Other works rule out
dimension-based characterizations for other learning problems.
\citet{LechnerBenDavid2024} show that no notion of dimension characterizes
the sample complexity of learning classes of distributions, and that no
characterization of their learnability satisfies a set of natural
requirements. They prove similar results for classification under
restricted marginal distributions and for classes of real-valued functions.
We use their notion of a finitary condition in the discussion after
\cref{def:intro-conditions}. \citet{BrukhimEtAl2025} show that no
dimension, in the sense of \citet{EMX2019}, characterizes bandit
learnability, even for finite classes. In contrast, our obstruction
depends on the size of the domain:
on every domain of cardinality less than $\aleph_\omega$, the conditions in
\cref{def:intro-conditions} characterize deterministic learnability.

\section{Proof of \cref{thm:independence}}
\label{sec:independence-proof}\label{sec:negative-proof}
    In this section, we prove \cref{thm:independence}, which we restate below.

    \MainIndependance*
\noindent
The following lemma verifies that $\mathcal M_A$, and
    in particular $\cH_\star$, satisfies the three conditions
    in \cref{def:intro-conditions}. Its proof appears in~\cref{sec:marker-conditions}.

    \begin{lemma} \label[lemma]{lem:marker-conditions}
        The class $\mathcal M_A$ in \cref{eq:example-class} has VC dimension two, uniform exterior separability, and proper singleton closures.
    \end{lemma}

\noindent
    By \cref{lem:marker-conditions} and
\citet[Theorem~4]{PositiveOnly}, $\cH_\star$ is learnable by a randomized learner in ZFC. This proves
part~(i) of \cref{thm:independence}.

\noindent
For part~(ii), we use the following two results on
the deterministic learnability of $\mathcal M_A$. Their proofs
appear in~\cref{sec:marker-small,sec:marker-large},
respectively.

    \begin{theorem} \label{thm:marker-small}
    The class $\mathcal M_A$ defined in~\cref{eq:example-class}
    is deterministically learnable whenever
    $|A|\leq\cc$, with sample complexity $m_{\mathcal M_A}(\varepsilon,\delta)=O((\varepsilon\delta)^{-3})$.
   
    \end{theorem}

    \begin{theorem}
    \label{thm:marker-large}
        The class $\mathcal M_A$ defined in~\cref{eq:example-class}
    is not deterministically learnable
    whenever $|A|\geq\beth_\omega$.
    \end{theorem}
\noindent
    We now combine \cref{thm:marker-small,thm:marker-large} to prove part~(ii) of
\cref{thm:independence}.
\begin{proof}[Proof of \cref{thm:independence}]
Let
\begin{equation} \label{eq:aleph-omega-marker}
    A_\star=\aleph_\omega \times\inbrace{0},\qquad B = \Nat \times\inbrace{1}, \qquad \mathcal H_\star=\mathcal M_{A_\star}, \qquad \cX_\star = \cX_{A_\star}\,.
\end{equation}

\noindent
Assume that ZFC is consistent.
Standard results in set theory provide a model $M_0$ of
ZFC satisfying GCH and a cardinal-preserving extension
$M_1$ in which $2^{\aleph_0}>\aleph_\omega$ \citep[Chapters~13 and~15]{Jech}.

\noindent
In $M_0$, GCH gives $|A_\star|=\aleph_\omega=\beth_\omega,$  and by \cref{thm:marker-large}, $\mathcal H_\star$ is not deterministically learnable. In $M_1$, the same set satisfies $|A_\star|=\aleph_\omega<2^{\aleph_0}.$
Therefore, by \cref{thm:marker-small}, $\mathcal{H}_\star$ is deterministically learnable.
Thus the assertion that $\cH_\star$ is deterministically learnable has opposite truth values
in the two models, so it is independent of ZFC.
\end{proof}

    \subsection{Proof of \cref{lem:marker-conditions}} \label{sec:marker-conditions}
    \begin{proof}[Proof of \cref{lem:marker-conditions}]
    Since every singleton belongs to the class, each singleton is its
    own closure, and singleton closures are proper. To verify UES, fix
    a nonempty finite realizable set $S$ and let $M=\ceil{1/\eta}$.
    If $|S|=1$,  repeat the singleton hypothesis $M$ times. If $|S|\ge2$
    and $S$ contains $b_j$, then $h_j$ is the unique consistent
    hypothesis, so we repeat it $M$ times. In the remaining case,
    $|S|\ge2$ and $S\subseteq A$. Its version space consists of all
    the $h_j$ and has intersection $A$. The list $h_1,\ldots,h_M$
    contains each point outside $A$ at most once, so each such point lies in at most a $1/M\leq\eta$ fraction of the list. This proves UES. Finally, $\VCdim(\mathcal{M}_A) = 2$.
    \end{proof}

    \subsection{Proof of \cref{thm:marker-small}} \label{sec:marker-small}
    \begin{proof}[Proof of \cref{thm:marker-small}]
Since \(|A|\leq\cc\), fix an injection \(\iota:A\to\R\).
Enumerate the rationals as \(r_1,r_2,\ldots\), and define
\[
 C_m=\{a\in A:\iota(a)<r_m\},\qquad
 v_m(S)=\sum_{x\in C_m}N_x(S)\,.
\]
Fix also the injective encoding $\pi(m,v)=2^{m-1}(2v-1)$ for $m,v\in\Nat.$
Thus an output index \(\pi(m,v)\) determines both \(m\) and \(v\).

\paragraph{Step 1 (Defining the learner).}
For a sample \(S\) of size \(n\ge27\), put \(t=n^{2/3}\) and apply
the following rules in order:
\begin{enumerate}
 \item If \(N_x(S)\geq n-t\) for some \(x\), return \(\{x\}\).
 \item Otherwise, if \(\supp(S)\cap B\neq\varnothing\), let
 \(j=\min\{i:b_i\in\supp(S)\}\) and return \(h_j\).
 \item Otherwise, let
 \[
  m(S)=\min\{m:t/2\leq v_m(S) \leq n - t/2\}\,,
 \]
 and return
 \[
  \mathcal A(S)=h_{\pi(m(S),v_{m(S)}(S))}\,.
 \]
\end{enumerate}
For \(n<27\), including the empty sample, return \(h_1\).

\noindent
The first rule selects at most one point because $n-t>n/2$.
To see that the third rule is defined, order the distinct sample points
$u_1,\ldots,u_q$ by their $\iota$-values, and let $i$ be the least index
with $\sum_{j=1}^{i}N_{u_j}(S)\geq t/2$. Since the first rule did not apply,
$N_{u_i}(S)<n-t$, while $\sum_{j=1}^{i-1}N_{u_j}(S)<t/2$.
Thus $t/2\leq\sum_{j=1}^{i}N_{u_j}(S)<n-t/2<n$, so $i<q$.
Any rational cut with $\iota(u_i)<r_m<\iota(u_{i+1})$ realizes this count, so the third rule is defined.
All choices are fixed functions of the multiset, and every output
belongs to \(\mathcal M_A\).

\paragraph{Step 2 (Bounding a fixed count).}
For \(1\leq v\leq n-1\), 
\begin{equation}\label{eq:MA-binomial}
 \sup_{p\in[0,1]}\Pr[\operatorname{Bin}(n,p)=v]
 =\binom nv(v/n)^v(1-v/n)^{n-v}
 \leq\sqrt{\frac{n}{v(n-v)}}\,.
\end{equation}
The maximum occurs at \(p=v/n\). 
The inequality follows from the Stirling bounds of \citet{Robbins1955}, which give the sharper bound \(e^{1/(12n)}\sqrt{n/[2\pi v(n-v)]}\).

\noindent
For a fixed $C\subseteq\cX_A$, let $E$ be the event that $\sum_{x\in C}N_x(S)=v$.
By \cref{asm:countable-support}, some countable $K\subseteq\cX_A$ has $\cD_+(K)=1$. The set $C$ need not be measurable, but $C\setminus K$ lies in the null set $\cX_A\setminus K$, so $C$ is measurable for the completion of $\cD_+$, with $\cD_+(C)=\cD_+(C\cap K)$. Hence this count has the binomial distribution with
parameters $n$ and $\cD_+(C)$, so
\begin{equation}\label{eq:MA-fixed-count}
 \Pr_{S\sim\cD_+^n}[E] = \Pr[\operatorname{Bin}(n,\cD_+(C))=v]
 \leq\sup_{p\in[0,1]}\Pr[\operatorname{Bin}(n,p)=v]\,.
\end{equation}

\noindent
In particular, if \(t/2\leq v\leq n-t/2\), then
\(v(n-v)\geq nt/4\), and \cref{eq:MA-binomial,eq:MA-fixed-count}
give
\begin{equation}\label{eq:MA-balanced-count}
 \Pr[E]\leq\frac2{\sqrt t}\,.
\end{equation}
\noindent
Now fix \(\ell\in\mathcal M_A\) and \(\cD\) with \(\cD(\ell)>0\).
If \(\ell=\{x\}\), all observations equal \(x\), so the first rule
returns \(\ell\) exactly. It remains to consider \(\ell=h_s\).

\paragraph{Step 3 (Bounding false positives).}
On a positive sample, the first rule returns a subset of \(\ell\).
The second rule returns \(\ell\), since \(b_s\) is the only marker
that can be observed. Thus false positives can arise only from the
third rule.

\noindent
Fix \(j\neq s\) with \(\cD(\{b_j\})>0\), and define the event $E_j$ by
\[
 E_j=\{S:\supp(S)\subseteq\ell,\ b_j\in\mathcal A(S)\}\,.
\]
On positive samples, \(E_j\) occurs exactly when the third rule returns \(h_j\).

\noindent
Write $j=\pi(m,v)$ for the unique pair $(m,v)\in\Nat^2$. If $v<t/2$ or $v>n-t/2$, the third
rule never returns $h_j$, so $E_j=\varnothing$. Otherwise, $E_j\subseteq\{S:v_m(S)=v\}$.
Applying \cref{eq:MA-balanced-count} to the fixed cut \(C_m\) gives
\(\Pr[E_j]\le2/\sqrt t\). This bound does not condition
on the sample-dependent choice \(m(S)\).
Since \(h_j\setminus h_s=\{b_j\}\) for \(j\neq s\),
\begin{equation}
 \E\,\cD(\mathcal A(S)\setminus\ell)
 =\sum_{\substack{j\neq s\\ \cD(\{b_j\})>0}}\cD(\{b_j\})\Pr[E_j]
 \leq\frac2{\sqrt t}\sum_{j\neq s}\cD(\{b_j\})
 \leq\frac2{\sqrt t}\,.
 \label{eq:MA-fp}
\end{equation}

\paragraph{Step 4 (Bounding false negatives).}
Each output of the learner misclassifies at most $t$ positive sample points. Standard PAC bounds give a universal constant $c_1$ such that
\begin{equation}
 \E\insquare{\cD(\ell\setminus\mathcal A(S))}
 \leq \E\insquare{\cD_+(\ell\setminus\mathcal A(S))}
 \leq t/n+c_1\sqrt{\frac{2\log(en)}{n}}\,.
 \label{eq:MA-fn}
\end{equation}
Combining \cref{eq:MA-fp} with \cref{eq:MA-fn} and using
\(t=n^{2/3}\) and $\log(en)\le3n^{1/3}$ gives
\[
 \E_{S\sim\cD_+^n}\err_{\cD}(\mathcal A(S),\ell)
 \leq \inparen{3+3c_1}n^{-1/3}\,.
\]
Therefore, for $n \geq \inparen{\frac{3+3c_1}{\varepsilon\delta}}^3$, Markov's inequality gives the desired bound
\[
 \Pr_{S\sim\cD_+^n}
   [\err_{\cD}(\mathcal A(S),\ell)>\varepsilon]
 \leq\delta\,.
\]
\end{proof}

    \subsection{Proof of \cref{thm:marker-large}} \label{sec:marker-large}
    
\begin{proof}[Proof of \cref{thm:marker-large}]
\noindent
Fix a deterministic proper learner $\mathcal A$ and a sample size
$n\ge1$. For $S\in[A]^n$, regard $S$ as a multiset in which each point occurs once, and color it by the learner's output:
\begin{equation}\label{eq:example-coloring}
 \chi(S)=
 \begin{cases}
  0,&\mathcal A(S)\text{ is a singleton},\\
  j,&\mathcal A(S)=h_j\quad(j\ge1)
 \end{cases}\,.
\end{equation}
This coloring uses at most countably many colors. By
\cref{lem:erdos-rado}, there is an infinite $F\subseteq A$
with a common color $j_\star\in\Nzero$, so
\begin{equation}\label{eq:example-homogeneous}
 \chi(S)=j_\star\qquad\text{for every }S\in[F]^n\,.
\end{equation}
\noindent
Choose a finite $F_0\subseteq F$ of size
$N\geq\max\{2,4\binom n2\}$. We will choose $\ell$ and
$\cD$ so that $\cD_+=\Unif(F_0)$.
Let $E$ be the event that the $n$ sampled points are distinct. Then
\begin{equation} \label{eq:example-distinct}
    \Pr[E]\ge1-\binom n2/N\ge3/4\,.
\end{equation}
On $E$, the sample is an $n$-element subset of $F$, and by~\mbox{\cref{eq:example-homogeneous}} its output has color $j_\star$.

\noindent We consider two cases.

\noindent
\emph{Case 1: $j_\star=0$.}
Take $\ell=h_1$ and $\cD=\Unif(F_0)$. Since $F_0\subseteq\ell$,
we have $\cD=\cD_+=\Unif(F_0)$. On $E$, the learner returns a
singleton $\{z\}$ with $\cD(\{z\})\leq\frac{1}{N}$, giving
\[
 \err_{\cD}(\{z\},\ell)
\ge1-\frac1N\geq\frac12\,.
\]

\noindent
\emph{Case 2: $j_\star\ge1$.}
Take $\ell=h_{j_\star+1}$ and
$\cD=\frac12\Unif(F_0)+\frac12\delta_{b_{j_\star}}$.
Since $F_0\subseteq A\subseteq\ell$ and $b_{j_\star}\notin\ell$, we have $\cD(\ell)=1/2$ and
$\cD_+=\Unif(F_0)$.
On $E$, the output is exactly $h_{j_\star}$. The point $b_{j_\star}$ belongs to
$h_{j_\star}\setminus\ell$ and has $\cD$-mass $1/2$, so
\[
 \err_{\cD}(h_{j_\star},\ell)
 \geq\cD(\{b_{j_\star}\})=\frac12\,.
\]
\noindent
In both cases, \cref{eq:example-distinct} gives the desired bound
\begin{equation}\label{eq:negative-error}
 \Pr_{S\sim\cD_+^n}
 \insquare{\err_{\cD}(\mathcal A(S),\ell)\geq\frac12}
 \geq\frac34\,.
\end{equation}
\end{proof}

\section{Proof of \cref{thm:no-finite-character}}
\label{sec:finite-character-proof}
In this section, we prove \cref{thm:no-finite-character}, which we restate below.

\MainNoFinite*
\noindent
We use the following invariance of bounded formulas \citep{FiniteCharacter}.

\begin{claim}[Invariance of bounded formulas]\label[claim]{clm:finite-character-models}
Let $M_1$ be a forcing extension of a model $M_0$ of ZFC, and let $\cX,\cH\in M_0$.
\begin{enumerate}
\item For every bounded formula $\varphi$, we have $M_0\models\varphi(\cX,\cH)$ if and
only if $M_1\models\varphi(\cX,\cH)$.
\item For every computable sequence $(\varphi_s)_{s\in\Nat}$ of bounded formulas, $M_0$
satisfies ``$\varphi_s$ holds for $(\cX,\cH)$ for every $s\in\Nat$'' if and only if $M_1$
does.
\end{enumerate}
\end{claim}

\begin{proof}
The sets $\cX$ and $\cH$, and every $h\in\cH$, have the same elements in $M_0$ and in
$M_1$, and the two models have the same membership relation on these elements. Hence both
models evaluate bounded formulas in the same structure $(\cX\cup\cH\cup\{\cX,\cH\};\in)$, which contains $\cX$ and $\cH$ as elements, and~(i) follows
by induction on $\varphi$. For~(ii), the two models also have the same natural numbers and
the same sequence $(\varphi_s)_{s\in\Nat}$, and the satisfaction relation of a fixed
set-sized structure is absolute between $M_0$ and $M_1$.
\end{proof}

\begin{proof}[Proof of \cref{thm:no-finite-character}]
Let $M_0\subseteq M_1$ be the models from the proof of \cref{thm:independence}: $M_0$
satisfies GCH, and $M_1$ is a cardinal-preserving forcing extension of $M_0$ with
$M_1\models2^{\aleph_0}>\aleph_\omega$. The domain $\cX_\star$ and the class $\cH_\star$
defined in \cref{eq:aleph-omega-marker} are the same sets in both models. By
\cref{thm:marker-large,thm:marker-small}, $\cH_\star$ is not deterministically learnable
in $M_0$ and is deterministically learnable in $M_1$.

Suppose first that a property $P$ of finite character characterizes deterministic
learnability, and let $\varphi$ be a bounded formula that ZFC proves equivalent to $P$.
Both models satisfy ZFC, so $M_0\models\neg\varphi(\cX_\star,\cH_\star)$ and
$M_1\models\varphi(\cX_\star,\cH_\star)$. This contradicts
\cref{clm:finite-character-models}~(i).

Now suppose that a dimension $\mathrm D$ of finite character characterizes deterministic
learnability, and let $(\varphi_s)_{s\in\Nat}$ be the associated sequence of bounded
formulas. Since $\cH_\star$ is not deterministically learnable in $M_0$, we have
$M_0\models\mathrm D(\cX_\star,\cH_\star)=\infty$. Equivalently, $M_0$ satisfies
``$\varphi_s$ holds for $(\cX_\star,\cH_\star)$ for every $s\in\Nat$''. By
\cref{clm:finite-character-models}~(ii), $M_1$ satisfies the same statement, so
$M_1\models\mathrm D(\cX_\star,\cH_\star)=\infty$. By the characterization, $\cH_\star$ is then not deterministically learnable in $M_1$, a contradiction.
\end{proof}

\section{Proof of \cref{thm:domain}}
\label{sec:pos-proof}
    In this section, we prove \cref{thm:domain}, which we restate below.
    \MainDomain*
\noindent
Part~(ii) follows from \cref{thm:marker-large}, and part~(iii) follows from parts~(i) and~(ii) under GCH. We prove part~(i) using the following three lemmas: one selects hypotheses, one converts a sample count into a probability bound, and one reconstructs finite samples from small messages.

    \noindent
\citet[Appendix~B.3]{PositiveOnly} bounded the
UES list size in~\cref{eq:ues} by
$M(\eta)=O\!\inparen{\frac{(d^\star+1)\log(e/\eta)}{\eta}}$.
Using this bound, the following lemma provides a simultaneous
selection property. For every point, it bounds the \emph{total number} of selected hypotheses that include the point outside their corresponding closures by $d^\star+1$, whereas UES bounds only the \emph{fraction} of such hypotheses in a single list.
Since finite VC dimension implies $d^\star<\infty$, this
gives a fixed finite bound independent of the number of
version spaces. Its proof appears in \cref{sec:selection-lem}.

    \begin{restatable}[Simultaneous finite selection]{lemma}{finiteSelectionLemma}
    \label[lemma]{lem:selection}
    Assume $d<\infty$ and UES, and set $r=d^\star+1$.
    For every finite list $V_1,\ldots,V_K$ of version spaces of
    nonempty finite realizable samples, there exist $h_i\in V_i$ with
    \begin{equation}\label{eq:overlap}
     \abs{\{i\in[K]:y\in h_i\setminus L(V_i)\}}\leq r
     \qquad(y\in\cX)\,.
    \end{equation}
    The version spaces may repeat.
    \end{restatable}
\noindent
The following lemma is similar in spirit to
\citet[Lemma~23]{PositiveOnly}, which they used to prove their
Theorem~7. Whereas that lemma bounds the probability of a fixed
multiset under an assumption on the distribution's largest point
mass, here we impose the condition $\max_xN_x(S)\leq n-t$ on
the multisets in the family $\cB$. This multiplicity bound
allows us to control the conditional probability of any
particular $S\in\cB$, given an enlarged sample
$T=S\uplus S'$, where $S'$ is an independent sample of the
same size, and $T$ is used only in the analysis. The proof of the lemma appears in \cref{sec:pool-lem}.

\begin{restatable}[From a finite-set count to a probability bound]{lemma}{pooledSampleLemma}
\label[lemma]{lem:pool}
Let $n\ge2$, $1\leq t\leq n/2$, and let $\cB$ be a family of
size-$n$ multisets such that for all $S \in \cB, \max_xN_x(S) \leq n-t.$ Suppose $b:\Nat\to[0,\infty)$ satisfies
\begin{align}
 |\{S\in\cB:\supp(S)\subseteq U\}|&\leq b(|U|)
 && \text{for every finite nonempty} ~U\subseteq\cX\,.
 \label{eq:pool-count-cond}
\end{align}
Then
\begin{equation}\label{eq:pool-bound}
 \Pr_{S\sim \cD_+^n}[S\in\cB]
 \leq\frac{12}{\sqrt t}\sup_{m\ge2}\frac{mb(m)}{2^m}\,.
\end{equation}
\end{restatable}

\noindent
A finite reconstruction map of size $k$ is a function
$\eta:[\cX]^{\leq k}\to[\cX]^{<\omega}$ such that every finite
$U\subseteq\cX$ has a subset $J$ with $|J|\leq k$ and
$U\subseteq\eta(J)$. The output is a finite set, whose size may depend on the input.

\noindent
The next lemma is the set-input form of \citet[Theorem~5.1]{FiniteCharacter}. For completeness, we include its proof in \cref{sec:reconstruction-lem}.

\begin{restatable}[Finite reconstruction below $\aleph_\omega$]{lemma}{finiteReconstructionLemma}
\label[lemma]{lem:reconstruction}
If $|\cX|\leq\aleph_{k-1}$ for some integer $k\ge1$, there is a
finite reconstruction map of size $k$. A finite domain admits size zero.
\end{restatable}
\noindent
We now prove \cref{thm:domain}~(i).

\begin{proof}[Proof of \cref{thm:domain}~(i)]
Assume the three conditions in \cref{def:intro-conditions}. Since $|\cX|<\aleph_\omega$,
there is an integer $k\ge1$ with $|\cX|\le\aleph_{k-1}$. Fix a reconstruction map $\eta$ of
size $k$, which exists by \cref{lem:reconstruction}, let $r=d^\star+1$, and fix a
well-order $\preceq$ of $[\cX]^{\leq k}$. With these choices fixed, we define a deterministic proper learner and bound its false-positive and false-negative errors separately.

\paragraph{Step 1 (Choosing finite tables).}
For each finite $U$, define the unique message
\begin{equation}\label{eq:rho}
 \rho(U)=\min_\preceq
 \{J\subseteq U:|J|\leq k,\ U\subseteq\eta(J)\}\,.
\end{equation}
The minimum exists because $\eta$ is a reconstruction map and $\preceq$ is a well-order. By definition, $\rho(U)\subseteq U$, $|\rho(U)|\leq k$,
and $U\subseteq\eta(\rho(U))$.
For $n\ge1$ and each message $J$, let
\begin{equation}\label{eq:message-table}
 \mathcal F_{n,J}=\{S:|S|=n,\ \cH_S\neq\varnothing,
                       \rho(\supp(S))=J\}\,.
\end{equation}
Every multiset in this family is supported on the finite set
$\eta(J)$, so the family is finite. For fixed $n$, the nonempty
families partition all realizable samples of size $n$.

\noindent
Apply \cref{lem:selection} to $(\cH_S:S\in\mathcal F_{n,J})$
to obtain $h_{n,J,S}\in\cH_S$ with
\begin{equation}\label{eq:table-property}
 \abs{\{S\in\mathcal F_{n,J}:
 y\in h_{n,J,S}\setminus\CLOS_{\cH}(S)\}}\leq r
 \qquad(y\in\cX)\,.
\end{equation}
For each realizable sample $S$ of size $n$, set
\begin{equation}\label{eq:gn}
 g_n(S)=h_{n,\rho(\supp(S)),S}\in\cH_S\,.
\end{equation}

\paragraph{Step 2 (Defining the learner).}
For $n\ge9$, put $t=n^{2/3}$, and for a realizable sample $S$, define
\begin{equation}\label{eq:learner}
 \mathcal A(S)=
 \begin{cases}
 \CLOS_{\cH}(\{x\}),&N_x(S)\geq n-t\text{ for some }x,\\[1mm]
 g_n(S),&\text{otherwise}
 \end{cases}\,.
\end{equation}
For $n<9$, let $\mathcal A(S)$ be an arbitrary concept in $\cH$.
Since $t<n/2$, at most one point satisfies $N_x(S)\geq n-t$,
which makes the first branch well-defined. Its output is proper
because $\cH$ has proper singleton closures. The second branch returns
a proper hypothesis consistent with $S$.

\paragraph{Step 3 (Counting samples with a false positive).}
For $y\in\cX$, let 
$\cB_{n,y}$ be the family of realizable multisets of size $n$ satisfying
\[
 \max_xN_x(S)<n-t,\qquad
 y\in g_n(S)\setminus\CLOS_{\cH}(S)\,.
\]
Fix a finite nonempty $U\subseteq\cX$ and put $m=|U|$. By \cref{eq:table-property},
\begin{equation}\label{eq:bad-count}
\begin{aligned}
 \abs{\{S\in\cB_{n,y}:\supp(S)\subseteq U\}}
 &=\sum_{\substack{J\subseteq U\\ |J|\leq k}}
   \abs{\{S\in\cB_{n,y}\cap\mathcal F_{n,J}:\supp(S)\subseteq U\}}\\
 &\leq r\Pi_k(m)\leq2^{d+1}\Pi_k(m)\,.
\end{aligned}
\end{equation}

\paragraph{Step 4 (Bounding false positives).}
Applying \cref{lem:pool} with $b(m)=2^{d + 1} \Pi_k(m)$ gives
\begin{equation}
 \Pr_{S\sim \cD_+^n}[S\in\cB_{n,y}]
 \leq\frac{12\,2^{d+1}\sup_{m\ge2}m\Pi_k(m)2^{-m}}{\sqrt t}\,.
 \label{eq:reconstruction-probability}
\end{equation}
For fixed $k$, $\Pi_k(m)$ is polynomial in $m$, so $\sup_{m\ge2}m\Pi_k(m)2^{-m}$
is finite. \cref{sec:aleph-omega-proof-calc}
gives the explicit bound
\begin{equation}
 12\,2^{d+1}\sup_{m\ge2}m\Pi_k(m)2^{-m}\leq48\,2^d(k+1)^2\,.
 \label{eq:reconstruction-constant}
\end{equation}
\noindent
Fix $\ell\in\cH$ and $\cD$ with $\cD(\ell)>0$.
For $y\notin\ell$, every positive sample satisfies
$y\notin\CLOS_{\cH}(S)$. The first branch of~\cref{eq:learner} returns a subset of $\ell$, since its selected point $x$ belongs
to $\ell$ and $\CLOS_{\cH}(\{x\})\subseteq\ell$.
A false positive at $y$ must therefore come from the second branch, in which case $S\in\cB_{n,y}$. Hence
\[
 \Pr[y\in\mathcal A(S)]
 \leq 48\,2^d(k+1)^2 n^{-1/3}\,.
\]

\noindent
Since $\cD$ is countably supported, summing this bound against the negative point masses of $\cD$ gives
\begin{equation}
 \E\,\insquare{\cD(\mathcal A(S)\setminus\ell)}
 \leq 48\,2^d(k+1)^2 n^{-1/3}\,.
 \label{eq:reconstruction-fp}
\end{equation}

\paragraph{Step 5 (Bounding false negatives).}
Each output of the learner misclassifies at most $t$ positive sample points. Standard PAC bounds give a universal constant $c_1$ such that
\begin{equation}
 \E\insquare{\cD(\ell\setminus\mathcal A(S))}
 \leq\E\insquare{\cD_+(\ell\setminus\mathcal A(S))}
 \leq t/n+c_1\sqrt{\frac{d\log(en)}{n}}\,.
\label{eq:reconstruction-fn}
\end{equation}
Combining \cref{eq:reconstruction-fp,eq:reconstruction-fn}, substituting $t=n^{2/3}$, and using $\log(en)\leq3n^{1/3}$ gives, for a universal constant $c$,
\[
    \E[\err_{\cD}(\mathcal A(S),\ell)]
 \leq \inparen{48\,2^d(k+1)^2+1+c_1\sqrt{3(d+1)}}n^{-1/3}
 \leq c\,2^d(k+1)^2n^{-1/3}\,.
\]
For $n \geq \inparen{\frac{c\,2^d(k+1)^2}{\varepsilon\delta}}^3$, Markov's inequality gives the desired bound
\[
 \Pr[\err_{\cD}(\mathcal A(S),\ell)>\varepsilon]
 \leq\frac{c\,2^d(k+1)^2}{\varepsilon n^{1/3}} \leq \delta\,.
\]
\end{proof}

\subsection{Proof of \cref{lem:selection}} \label{sec:selection-lem}
\begin{proof}[Proof of \cref{lem:selection}]
    If $K\leq r$, any choice of one hypothesis from each $V_i$ works.
Assume $K>r$. For each $i$, take a UES list
$f_{i,1},\ldots,f_{i,M}\in V_i$ at scale $\eta$.
\citet[Appendix~B.3]{PositiveOnly} give the list-size bound
\begin{equation}\label{eq:quant-ues}
 M=O\!\inparen{\frac{(d^\star+1)\log(e/\eta)}{\eta}}\,.
\end{equation}
Choose $J_1,\ldots,J_K$ independently and uniformly from $[M]$.

\noindent
For a fixed $y$, each event $y\in f_{i, J_i}\setminus L(V_i)$  has probability at most $\eta$. The events are independent across $i$. Thus
\begin{equation} \label{eq:overlap-y}
 \Pr\!\insquare{\abs{\{i:y\in f_{i, J_i} \setminus L(V_i)\}}\geq r+1}
 \leq\binom K{r+1}\eta^{r+1}\,.
\end{equation}
\noindent
Let $E_y$ be the event that $y$ belongs to at least $r+1$
of the sets $f_{i, J_i}\setminus L(V_i)$. To bound the number of distinct
events, consider points $y,z$ with the same memberships in every
candidate hypothesis $f_{i,j}$ and every closure $L(V_i)$.
For any choice of $J_1,\ldots,J_K$, the same indices contribute
to the overlap count at $y$ and at $z$, so $E_y=E_z$.
Thus the events are determined by these finite membership patterns,
even when $\cX$ is infinite.

\noindent
There are at most $KM$ candidate hypotheses. Since
$\VCdim(\cH^\star)=d^\star$, Sauer's lemma on $\cH^\star$ bounds the number
of membership patterns on these hypotheses by $(KM+1)^{d^\star}$.
Membership in $L(V_1),\ldots,L(V_K)$ forms a binary vector
of length $K$, with at most $2^K$ possible values.

\noindent
There are therefore at most $2^K(KM+1)^{d^\star}$ distinct bad
events. Choose one representative point for each realized pair
of patterns. A union bound using~\cref{eq:overlap-y} gives
\[
\begin{aligned}
\Pr\!\insquare{\bigcup_{y\in\cX}E_y}
&\leq 2^K(KM+1)^{d^\star}\binom K{r+1}\eta^{r+1}
=O_{K,d^\star}\!\inparen{\eta^2[\log(e/\eta)]^{d^\star}}\,.
\end{aligned}
\]
\noindent
The last equality uses~\cref{eq:quant-ues} and $r+1=d^\star+2$.
For fixed $K,d^\star$, the bound tends to zero with $\eta$.
Hence a tuple with no bad event exists. For such a tuple, the hypotheses $h_i=f_{i,J_i}$ satisfy \cref{eq:overlap}.
\end{proof}

\subsection{Proof of \cref{lem:pool}} \label{sec:pool-lem}
\noindent To prove \cref{lem:pool}, we first establish the following binomial bounds. We will use them to bound the fraction of index subsets that produce each candidate sample.

\begin{lemma} \label[lemma]{lem:binom-calc}
    For $n,N\in\Nat$ and $s\leq N$, we have
    \[
 2^{-N}\binom Ns\leq\min\!\inbrace{\frac12,\sqrt{\frac1N}}
 \quad(N\ge1)\qquad\text{and}\qquad
 2^{-2n}\binom{2n}{n}\geq\frac1{2\sqrt n}\,.
  \]
\end{lemma}
\begin{proof}[Proof of \cref{lem:binom-calc}]
    For $j\ge0$, define
$b_j\coloneqq4^{-j}\binom{2j}{j}=\frac{(2j)!}{4^j(j!)^2}$.
Then $b_0=1$ and $b_1=1/2$, and cancelling factorials gives
$\frac{b_{j+1}}{b_j}=\frac{2j+1}{2j+2}$.
For every $j\geq0$,
\[
\sqrt{\frac{j}{j + 1}} \leq \frac{2j+ 1}{2j + 2} \leq \sqrt{\frac{j + 1}{j + 2}}\,.
\]
\noindent
Iterating these inequalities gives
$\frac{1}{2\sqrt{n}}\leq b_n\leq\frac{1}{\sqrt{2(n+1)}}$,
which proves the second bound. For the first bound, if $N$ is even,
then $2^{-N}\binom Ns\leq b_{N/2}\leq\sqrt{\frac{1}{N}}$.
If $N=2j+1$ is odd, then
\[
2^{-N}\binom Ns \leq  2^{-(2j+1)}\binom{2j+1}{j}
=\frac{2j+1}{2j+2}\,b_j
 =b_{j+1}\leq \frac{1}{\sqrt{2(j + 2)}} \leq \frac{1}{\sqrt{N}}\,.
\]
\end{proof}

\begin{proof}[Proof of \cref{lem:pool}]
\noindent Draw $x_1,\ldots,x_{2n}$ independently from $\cD_+$, and let
$S_I$ be the multiset that contains $x_i$ once for each $i\in I$, for $I\subseteq[2n]$ with $|I|=n$.

\noindent
Each $S_I$ has the law of $n$ independent observations from $\cD_+$, so
\begin{equation}\label{eq:pool-index-average}
 \Pr_{S \sim \cD_+^{n}}[S\in\cB]
 =\E \!\insquare{\frac1{\binom{2n}{n}} \sum_{\substack{I\subseteq[2n]\\|I|=n}}\1\{S_I\in\cB\}} \,.
\end{equation}

\noindent
We bound this average for each fixed pool of $2n$ observations.
Let $T$ be their multiset. If $T$ has no size-$n$ submultiset in
$\cB$, the average is zero. Otherwise, write
$\supp(T)=\{u_1,\ldots,u_m\}$, where $u_i$ has multiplicity $N_i$ in $T$ and $N_1\geq\cdots\geq N_m\ge1$. A sample in $\mathcal B$ contains
at most $n-t$ copies of $u_1$, and its complement in $T$ has size
$n$, so $N_1\le2n-t$. The presence of such a sample also gives
$m\geq2$. It follows that
\begin{equation}\label{eq:pool-two-counts}
 N_1\geq\frac{2n}{m},\qquad
 N_2\geq\frac{2n-N_1}{m-1}\geq\frac{t}{m-1}\,.
\end{equation}
For a size-$n$ submultiset $R\subseteq T$ with multiplicity $s_i$
at $u_i$, exactly $\prod_i\binom{N_i}{s_i}$ index subsets produce
$S_I=R$. The fraction of all index subsets producing $R$ is therefore
\begin{equation}\label{eq:pool-index-fraction}
 w_T(R)=\frac{\prod_{i=1}^m\binom{N_i}{s_i}}{\binom{2n}{n}}\,.
\end{equation}

\noindent
The elementary central-binomial bounds proved in
\cref{lem:binom-calc} give
\begin{equation}\label{eq:binomial}
 2^{-N}\binom Ns\leq\min\!\inbrace{\frac12,\sqrt{\frac1N}}
 \quad(N\ge1)\qquad\text{and}\qquad
 2^{-2n}\binom{2n}{n}\geq\frac1{2\sqrt n}\,.
\end{equation}

\noindent Using the inverse-square-root bounds for $N_1,N_2$ and $1/2$ for
the other factors gives
\begin{equation}
 w_T(R)\leq\frac{2\sqrt n}{\sqrt{N_1N_2}}2^{-(m-2)}
 \leq 2\sqrt{\frac{m(m-1)}{2t}}2^{-(m-2)}
 <\frac{12m\,2^{-m}}{\sqrt t}\,.
\label{eq:pool-one-candidate}
\end{equation}
At most $b(m)$ members of $\cB$ are supported on $\supp(T)$,
so there are at most this many candidates among the submultisets
of $T$. Summing~\cref{eq:pool-one-candidate} over them bounds
the average in~\cref{eq:pool-index-average} by
$12m b(m)2^{-m}/\sqrt t$. Taking the supremum over $m\ge2$
and averaging over the pool proves~\cref{eq:pool-bound}.
\end{proof}

\subsection{Proof of \cref{lem:reconstruction}} \label{sec:reconstruction-lem}

\begin{proof}[Proof of \cref{lem:reconstruction}]
If the domain is finite, set $\eta(\varnothing)=\cX$.
The empty message then reconstructs every finite subset, giving
size zero. To allow a larger message size, define $\eta$ arbitrarily on the remaining inputs.

\noindent
For a countably infinite domain and $k=1$, enumerate
$\cX=\{x_1,x_2,\ldots\}$. Set $\eta(\varnothing)=\varnothing$
and define
\[
 \eta(\{x_j\})=\{x_1,\ldots,x_j\}\,.
\]
\noindent
For a nonempty $U=\{x_{i_1},\ldots,x_{i_m}\}\subseteq\cX$, the point with
largest index gives the message, since
$U\subseteq\eta(\{x_{\max\{i_1,\ldots,i_m\}}\})$.

\noindent
Now suppose $k\ge2$ and assume the result for $k-1$.
Well-order $\cX$ by $\prec$ with order type the initial ordinal
of cardinality $|\cX|$. For each $x\in\cX$, the proper initial
segment $\cX_{<x}\coloneqq\{z\in\cX:z\prec x\}$ has cardinality at most
$\aleph_{k-2}$.
The induction hypothesis therefore gives a reconstruction map
$\eta_x:[\cX_{<x}]^{\leq k-1}\longrightarrow[\cX_{<x}]^{<\omega}$
for each $x$. Set $\eta(\varnothing)=\varnothing$.
For a nonempty $J\in[\cX]^{\leq k}$, let $x=\max_\prec J$
and define
\begin{equation}
 \eta(J)=\{x\}\cup\eta_x(J\setminus\{x\})\,.
 \label{eq:reconstruction-recursion}
\end{equation}
The output is finite. To check reconstruction, take a nonempty
finite $U$ and let $x=\max_\prec U$. The induction hypothesis
supplies $J'\subseteq U\setminus\{x\}$ of size at most $k-1$
such that $U\setminus\{x\}\subseteq\eta_x(J')$.
The message $J=J'\cup\{x\}\subseteq U$ then has $|J|\leq k$,
and~\cref{eq:reconstruction-recursion} gives $U\subseteq\eta(J)$,
completing the induction.
\end{proof}

\subsection{Technical Result Used in the Proof of \cref{thm:domain}} \label{sec:aleph-omega-proof-calc}

\begin{lemma} \label[lemma]{lem:rec-calc}
    For every $k\in\Nat$, we have
    \[
    \sup_{m\ge2}m\Pi_k(m)2^{-m} \leq 2(k+1)^2\,.
    \]
\end{lemma}
\begin{proof}[Proof of \cref{lem:rec-calc}]
For $0<z<1$ and $j\ge0$,
\[
 \sum_{m=0}^{\infty}\binom mj z^m
 =\frac{z^j}{(1-z)^{j+1}}\,.
\]
Differentiating with respect to $z$ gives
\[
 \sum_{m=0}^{\infty}m\binom mj z^{m-1}
= \inparen{\frac{j}{z} + \frac{j+1}{1 - z}} \frac{z^j}{(1-z)^{j+1}}\,.
\]
At $z=1/2$, we obtain
\[
 \sum_{m=0}^{\infty}m\binom mj2^{-m}=4j+2\,.
\]
All summands are nonnegative, so
\begin{align*}
 \sup_{m\geq2}m\Pi_k(m)2^{-m}
 &\leq\sum_{m=0}^{\infty}m\Pi_k(m)2^{-m}\\
 &=\sum_{j=0}^k\sum_{m=0}^{\infty}m\binom mj2^{-m}\\
 &=\sum_{j=0}^k(4j+2)\\
 &=2(k+1)^2\,.\qedhere
\end{align*}
\end{proof}

\section{Proof of \cref{thm:uniform-cardinal-condition}}
\label{sec:uniform-cardinal-proof}

\noindent
In this section, we prove \cref{thm:uniform-cardinal-condition}, which we restate below.

\MainCardCondition*
\begin{proof}[Proof of \cref{thm:uniform-cardinal-condition}]
We can express~\cref{eq:uniform-cardinal-condition} using a single
infinite cardinal $\kappa<\beth_\omega$ satisfying
\begin{equation}\label{eq:common-kappa}
 0<|\cH_A|\leq\kappa,\qquad
 \text{ for all} ~h\in (\cH\setminus\cH_A), ~|h\cap A|\leq\kappa\,.
\end{equation}

\noindent
Fix a deterministic proper learner $\mathcal A$ and a sample size $n\ge1$.
Choose $\kappa$ satisfying~\cref{eq:common-kappa}, and enumerate
\[
 \cH_A=\{h_\xi:\xi<\theta\},\qquad 0<\theta\leq\kappa\,,
\]
without repetitions, identifying cardinals with their initial
ordinals when they serve as indices.

\paragraph{Step 1 (Coloring samples by the learner's output).}
For each $S\in[A]^n$, define
\begin{equation}\label{eq:output-type}
 c(S)=
 \begin{cases}
  \xi,&\mathcal A(S)=h_\xi\in\cH_A,\\
  *,&\mathcal A(S)\notin\cH_A
 \end{cases}\,.
\end{equation}
Here $*$ is distinct from every index $\xi<\theta$.
The index $\xi$ identifies the exact output $h_\xi$. The symbol
$*$ groups the outputs that do not contain $A$, which can vary
with the sample. To control those outputs, we also record their
predictions on an additional point.

\noindent
Fix a well-order $\prec$ of $A$. For
$T=\{x_1\prec\cdots\prec x_{n+1}\}\in[A]^{n+1}$, let
$S_i=T\setminus\{x_i\}$ and set
\begin{equation}\label{eq:deletion-color}
 \chi(T)=
 \inparen{c(S_i),\ \1\{x_i\in\mathcal A(S_i)\}}_{i=1}^{n+1}\,.
\end{equation}
Coordinate $i$ records the output type after deleting $x_i$ and
whether the resulting hypothesis contains $x_i$. The well-order
specifies the coordinate positions, while the learner receives only
the unordered multiset $S_i$.

\noindent
The colors $\chi(T)$ belong to
$((\theta\cup\{*\})\times\{0,1\})^{n+1}$. Since $\kappa$ is
infinite and $n$ is finite, this set has cardinality at most
\[
 (\kappa\cdot2)^{n+1}=\kappa\,.
\]
Thus the number of colors is at most
$\kappa<\beth_\omega$.

\paragraph{Step 2 (Finding a homogeneous set).}

We use the partition theorem (\cref{thm:erdos-rado}) with $\lambda=\kappa^+$.
A set of size $\lambda$ is then too large
to lie in any intersection $h \cap A$ for $h\notin\cH_A$ (each has cardinality at most $\kappa$ by~\cref{eq:common-kappa}). To apply the partition theorem for
$[A]^{n+1}$, we check its cardinality requirement. Since $\kappa<\beth_\omega$,
there is a finite $s$ with $\kappa\leq\beth_s$, and hence
\[
 \lambda=\kappa^+\leq\beth_s^+\leq\beth_{s+1}<\beth_\omega\,.
\]
Choose a finite $s_0$ with $\lambda\leq\beth_{s_0}$.
Since taking power sets preserves cardinal inequalities, 
\begin{equation}\label{eq:uniform-er-size}
 \exp_n(\lambda)^+
 \leq\beth_{s_0+n}^+
 \leq\beth_{s_0+n+1}
 <\beth_\omega
 \leq|A|\,.
\end{equation}
The coloring uses at most $\kappa\leq\lambda$ colors, so the
partition theorem applies and gives a homogeneous subset of $A$ of cardinality $\lambda^+$. Its inherited well-order has order type at least
the initial ordinal $\lambda^+$. Restrict to its first $\lambda^+$
elements, which preserves homogeneity, and write
\[
 F=\{f_\alpha:\alpha<\lambda^+\},\qquad
 \alpha<\beta\ \Longrightarrow\ f_\alpha\prec f_\beta\,.
\]
Let its common color be
\begin{equation}\label{eq:common-deletion-color}
 \chi(T)=((c_1,e_1),\ldots,(c_{n+1},e_{n+1}))
 \qquad(\text{for all} ~T\in[F]^{n+1})\,.
\end{equation}
The vector is independent of $T$, although its coordinates may
differ from one another. We next show that all the coordinates $c_i$ agree.

\paragraph{Step 3 (Fixing the output type).}
For $i=1,\ldots,n$, put
\[
 p_i=f_{\lambda\cdot i},\qquad P=\{p_1,\ldots,p_n\}\,.
\]
Here $\lambda\cdot i$ denotes ordinal multiplication: the order
type of $i$ consecutive copies of $\lambda$. For every positive
finite $i$, the ordinal $\lambda\cdot i$ has cardinality $\lambda$, so
$\lambda\cdot i<\lambda^+$. The points $p_i$ are therefore defined
and occur in increasing order.

\noindent
These points divide the remaining points of $F$ into $n+1$ gaps:
\[
 \begin{aligned}
 I_1&=\{z\in F:z\prec p_1\},\\
 I_i&=\{z\in F:p_{i-1}\prec z\prec p_i\}
                   &&(2\leq i\leq n),\\
 I_{n+1}&=\{z\in F:p_n\prec z\}\,.
 \end{aligned}
\]
Each gap has cardinality at least $\lambda$. The first contains
all points indexed by $\alpha<\lambda$. Each intermediate gap contains the block between $\lambda\cdot(i-1)$ and
$\lambda\cdot i$, with its endpoint removed. The last gap contains
the analogous block between $\lambda\cdot n$ and
$\lambda\cdot(n+1)$, both of which lie below $\lambda^+$.
Every block has cardinality $\lambda$, and removing finitely many
endpoints preserves that cardinality.

\noindent
Fix $i\in[n+1]$ and $z\in I_i$. The point $z$ has position $i$
in the increasing enumeration of $P\cup\{z\}$, and deleting it
leaves $P$. Comparing the $i$th coordinates
in~\cref{eq:deletion-color} and~\cref{eq:common-deletion-color}
therefore gives
\begin{equation}\label{eq:gap-colors}
 c_i=c(P),\qquad e_i=\1\{z\in\mathcal A(P)\}
 \qquad(z\in I_i)\,.
\end{equation}
Since every gap is nonempty, each $c_i$ equals $c(P)$.
Write $c$ for the common value $c_1=\cdots=c_{n+1}$.

\noindent
Now take any $S\in[F]^n$ and any $z\in F\setminus S$. If $z$ has
position $i$ in the increasing enumeration of $S\cup\{z\}$, the
coordinate that deletes $z$ gives
\begin{equation}\label{eq:all-samples-one-type}
 c(S)=c_i=c,\qquad
 \1\{z\in\mathcal A(S)\}=e_i\,.
\end{equation}
The coloring is homogeneous on $(n+1)$-element subsets of $F$,
so every $n$-element sample from $F$ has the same output type.

\paragraph{Step 4 (The learner's behavior on $F$).}
If $c=\xi<\theta$, the definition of $c(S)$ and
\cref{eq:all-samples-one-type} imply
\begin{equation}\label{eq:constant-large-output}
 \mathcal A(S)=h_\xi\qquad\text{for every }S\in[F]^n\,.
\end{equation}

\noindent
Suppose instead that $c=*$. Since $c(P)=*$, properness gives
$\mathcal A(P)\in\cH\setminus\cH_A$. The uniform intersection
bound implies
\[
 |\mathcal A(P)\cap A|\leq\kappa<\lambda\,.
\]
If $e_i=1$ for some $i$, then~\cref{eq:gap-colors} places the
entire gap $I_i$ in $\mathcal A(P)\cap A$, giving
\[
 \lambda\leq |I_i|\leq|\mathcal A(P)\cap A|\leq\kappa\,,
\]
a contradiction. Therefore $e_i=0$ for every $i$.

\noindent
For any $S\in[F]^n$ and $z\in F\setminus S$, the second equality
in~\cref{eq:all-samples-one-type} now gives $z\notin\mathcal A(S)$.
Equivalently,
\begin{equation}\label{eq:small-output-on-F}
 \mathcal A(S)\cap F\subseteq S\qquad(S\in[F]^n)\,.
\end{equation}
The output includes no point of $F$ outside the sample. Whether
it contains all of $S$ is unrestricted, since the argument makes
no consistency assumption on the learner.

\paragraph{Step 5 (Forcing constant error).}
Choose a finite $F_0\subseteq F$ of size
$N\geq\max\{2n,4\binom n2\}$. We will choose $\ell$ and
$\cD$ so that $\cD_+=\Unif(F_0)$.
Let $E$ be the event that the $n$ sampled points are distinct. Then
\begin{equation} \label{eq:uniform-distinct}
    \Pr[E]\ge1-\binom n2/N\ge3/4\,.
\end{equation}
On $E$, the sample lies in $[F_0]^n\subseteq[F]^n$, so the
conclusions of Step 4 apply.

\noindent
If $c=*$, choose any $\ell\in\cH_A$, which exists by realizability,
and set $\cD=\Unif(F_0)$. Since $F_0\subseteq A\subseteq\ell$, we have
$\cD(\ell)=1$ and thus $\cD_+ = \cD$. On $E$,
\cref{eq:small-output-on-F} gives
\[
 \mathcal A(S)\cap F_0\subseteq S,\qquad
 |\mathcal A(S)\cap F_0|\leq n\,.
\]
Since $\cD$ is uniform on $F_0$ and $N\ge2n$,
\[
 \err_{\cD}(\mathcal A(S),\ell)
 =1-\frac{|\mathcal A(S)\cap F_0|}{N}
 \ge1-\frac nN\geq\frac12\,.
\]
\noindent
If $c=\xi<\theta$, let $L=\CLOS_{\cH}(A)$.
Every member of $\cH_A$ contains $L$, so $L\subseteq h_\xi$.
The inclusion is strict because $L\notin\cH$. Choose
$z\in h_\xi\setminus L$. By the definition
$L=\bigcap_{h\in\cH_A}h$, some $\ell\in\cH_A$ excludes $z$.
Since $A\subseteq L$, the point $z$ also lies outside $A$ and $F_0$.

\noindent
Set $\cD=\frac12\Unif(F_0)+\frac12\delta_z$.
Since $F_0\subseteq A\subseteq\ell$ and $z\notin\ell$, we have $\cD_+ = \Unif(F_0)$. On $E$,
\cref{eq:constant-large-output} makes the output $h_\xi$, which
contains the negative point $z$. Therefore
\[
 \err_{\cD}(\mathcal A(S),\ell)
 \geq\cD(\{z\})=\frac12\,.
\]
\noindent
In both cases, the error is at least $1/2$ on $E$.
Combining this with~\cref{eq:uniform-distinct} gives
\cref{eq:uniform-cardinal-error}. Because $\mathcal A$ and $n$
were arbitrary, the PAC guarantee fails for every deterministic
proper rule at $\varepsilon=\delta=1/4$.
\end{proof}

\end{document}